\documentclass[10pt,journal,compsoc]{IEEEtran}

\usepackage[OT1]{fontenc} 
\usepackage[utf8]{inputenc}
\usepackage{amsmath,amsfonts,amssymb,amsthm}
\usepackage{graphicx}
\usepackage{color}
\usepackage{tabularx}
\usepackage[table]{xcolor}
 \usepackage{booktabs}
\usepackage{caption}
\usepackage{subcaption}
\usepackage{algorithm}
\usepackage[noend]{algpseudocode}
\usepackage{xspace}
\usepackage{enumitem} 
\usepackage{amssymb}  
\usepackage{ifthen} 
\usepackage[nocompress]{cite}
\usepackage{array}
\usepackage{multirow}
\usepackage[pagebackref=true,breaklinks=true,letterpaper=true,colorlinks,bookmarks=false]{hyperref}
\usepackage{pifont}
\usepackage{soul}
\usepackage{cleveref}
\newcommand\labelLastPage{%
  \edef\@currentlabel{\thepage}%
  \label{LastPage}%
}

\ifCLASSOPTIONcompsoc
  \usepackage[nocompress]{cite}
\else
  \usepackage{cite}
\fi
\usepackage{amssymb}
\usepackage{booktabs}
\usepackage{multirow}
\usepackage{graphicx}
\usepackage{amsmath}

\ifCLASSINFOpdf
\else
\fi
\begin{document}
%
\title{DebSDF: Delving into the Details and Bias of Neural Indoor Scene Reconstruction}

\author{Yuting~Xiao*,
        Jingwei~Xu*,
        Zehao~Yu,
        Shenghua Gao
\IEEEcompsocitemizethanks{\IEEEcompsocthanksitem Yuting Xiao and Jingwei Xu contributed equally to this work;
\IEEEcompsocthanksitem Corresponding Author: Shenghua Gao;\protect\\
E-mail: gaosh@hku.hk
\IEEEcompsocthanksitem Yuting Xiao, Jingwei Xu, and Shenghua Gao are with the School of Information Science and Technology, ShanghaiTech University, Shanghai 201210, China;
\IEEEcompsocthanksitem Zehao Yu is with the Department of Computer Science, University of Tübingen, 72076, Germany;
\IEEEcompsocthanksitem Shenghua Gao is with the University of Hong Kong, Hong Kong SAR, China;

}
}

%
%

\markboth{Journal of \LaTeX\ Class Files,~Vol.~14, No.~8, August~2015}%
{Shell \MakeLowercase{\textit{et al.}}: Bare Demo of IEEEtran.cls for Computer Society Journals}

\IEEEtitleabstractindextext{%

\begin{abstract}
In recent years, the neural implicit surface has emerged as a powerful representation for multi-view surface reconstruction due to its simplicity and state-of-the-art performance. However, reconstructing smooth and detailed surfaces in indoor scenes from multi-view images presents unique challenges. Indoor scenes typically contain large texture-less regions, making the photometric loss unreliable for optimizing the implicit surface. Previous work utilizes monocular geometry priors to improve the reconstruction in indoor scenes. However, monocular priors often contain substantial errors in thin structure regions due to domain gaps and the inherent inconsistencies when derived independently from different views. This paper presents \textbf{DebSDF} to address these challenges, focusing on the utilization of uncertainty in monocular priors and the bias in SDF-based volume rendering. We propose an uncertainty modeling technique that associates larger uncertainties with larger errors in the monocular priors. High-uncertainty priors are then excluded from optimization to prevent bias. This uncertainty measure also informs an importance-guided ray sampling and adaptive smoothness regularization, enhancing the learning of fine structures. We further introduce a bias-aware signed distance function to density transformation that takes into account the curvature and the angle between the view direction and the SDF normals to reconstruct fine details better. Our approach has been validated through extensive experiments on several challenging datasets, demonstrating improved qualitative and quantitative results in reconstructing thin structures in indoor scenes, thereby outperforming previous work. The source code and more visualizations can be found in \url{https://davidxu-jj.github.io/pubs/DebSDF/}.

\end{abstract}


}

\maketitle

\IEEEdisplaynontitleabstractindextext

\IEEEpeerreviewmaketitle

\begin{figure*}[t]
\centering
\includegraphics[width=0.9\textwidth]{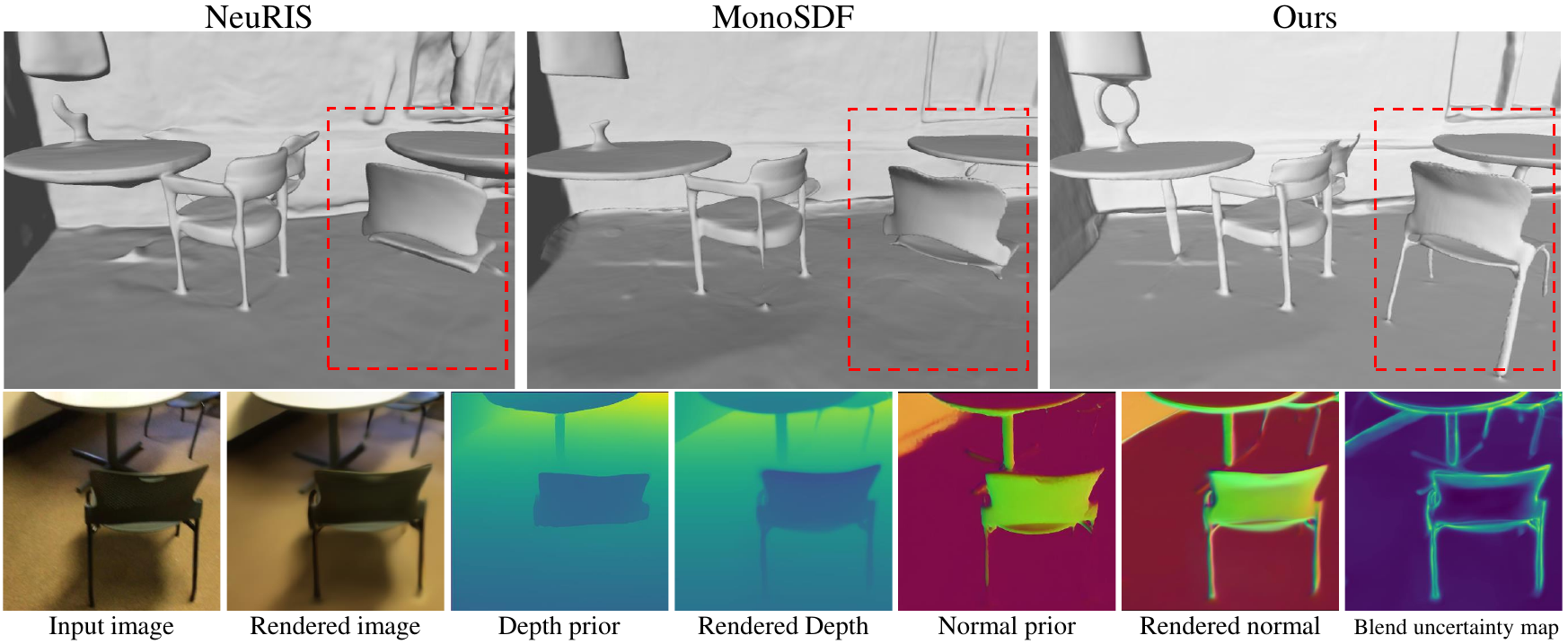}
\caption{We can observe that our method can reconstruct the indoor scene with more detailed structures, such as the chair legs and bracket of the desk lamp. Previous works such as MonoSDF \cite{yu2022monosdf}, which is based on the VolSDF \cite{yariv2021volume}, can not reconstruct the thin and detailed surface due to the inaccurate geometry prior at these regions. 
Our method can accurately generate the uncertainty map, which can localize the inaccurate priors and reduce the bias in SDF-based rendering with a proposed bias-aware SDF to density transformation approach so that our method can reconstruct the indoor scene significantly better than previous works. 
}
\label{fig:teaser}
\end{figure*}

\IEEEraisesectionheading{\section{Introduction}\label{sec:introduction}}

\IEEEPARstart{S}{urface} reconstruction from multiple calibrated RGB images is a long-standing goal in computer vision and graphics with various applications ranging from robotics to virtual reality. Traditional methods tackle this problem with a multi-step pipeline. It first estimates dense depth maps for each RGB image by the multi-view stereo, fuses the depth maps to unstructured point clouds, and then converts it to a triangle mesh with Poisson surface reconstruction \cite{kazhdan2013screened}. 
Recently, the neural implicit surface \cite{yariv2021volume, wang2021neus, oechsle2021unisurf, mildenhall2021nerf, niemeyer2020differentiable} has emerged as a powerful representation for multi-view surface reconstruction due to its simplicity. 
The key ideas are the utilization of coordinate-based networks for scene representation that map 3D coordinates to different scene properties such as signed distance and the use of differentiable volume rendering techniques that project the implicit surface to 2D observations such as RGBs, depths, and normals. The neural implicit surface can be optimized with multi-view images in an end-to-end manner and can be further converted to triangle mesh easily with the marching cubes algorithm at arbitrary resolution.

Despite the impressive reconstruction performance that has been shown for object-centric scenes with implicit surfaces, it is still challenging to reconstruct smooth and detailed surfaces from multi-view images for indoor scenes. First, multi-view reconstruction is an inherently under-constrained problem as there exists an infinite number of plausible implicit surfaces that match the input images after rendering. Second, indoor scenes usually contain large texture-less regions and the photometric loss used for optimizing the implicit surface is unreliable. One possible way to circumvent these challenges is to leverage the priors about the indoor scenes. For example, by assuming a Manhattan world \cite{coughlan1999manhattan} where the wall and floor regions are orthogonal, Manhattan-SDF \cite{guo2022neural} achieves better performance than methods that directly optimize SDF from multi-view images. MonoSDF~\cite{yu2022monosdf} and NeuRIS \cite{wang2022neuris} further improve the reconstruction quality by incorporating large pre-trained models to infer geometric prior from a single view, e.g.\ monocular depth and normal maps. 

While the overall structures such as walls and floors of indoor scenes can be faithfully reconstructed, these methods still struggle to recover fine details, e.g.\ chair legs and bracket of the desk, as shown in Fig~\ref{fig:teaser}. These failures result from several factors: 1) Monocular priors have large errors in these regions due to domain gaps between the training data used for the pre-trained model and the scenes we want to reconstruct. 2) As the monocular priors of each input image are predicted independently, they are unlikely to be multi-view consistent, especially in the thin and detailed structure regions. 3) Thin structure regions occupy a small area in the input images. Existing methods sample training rays uniformly across all training images, resulting in low sampling probabilities of these regions in comparison to walls and floors. 4) Smoothness regularization is applied uniformly to the whole space, which suppresses the learning of thin structure surfaces. These combined factors together make fine and detailed surfaces hard to form. 

In this paper, we address these challenges of utilizing monocular geometric priors in a unified perspective. Our key insight is based on the assumption that a prior is correct if it is consistent with other priors, e.g.\ consistent with other views or other modalities (depth normal consistency in our case). While manually detecting this consistency or inconsistency can be challenging, we observe that the optimization of implicit surface implicitly aggregates information/priors from multi-views and multi-modality (depth and normal). Geometric priors that are consistent with others will have low errors, while priors that deviate from others will have large errors in comparison to learned implicit surfaces. Therefore, we achieve this with uncertainty modeling where large uncertainty reflects the large error of the priors. 
we introduce the uncertainty into the monocular prior guided optimization to avoid the effect of wrong priors in surface reconstruction. We observe that the filtered regions usually correspond to thin structure regions, as shown in the \textit{Importance map} in Fig~\ref{fig:teaser}. Therefore, we propose an importance-guided ray sampling that samples more rays from these regions and an adaptive smoothness regularization strategy that applies smaller regularization on these regions to facilitate the learning of fine structures.

Further, we find that the commonly used SDF to density transformation has a non-negligible bias when a ray is casting around the surface where multiple peaks exist in the volume rendering weights. Optimization with such an ambiguous formula suppresses the reconstruction of foreground objects, e.g.\ thin structures in indoor scenes. We propose to transform the SDF to density with the curvature radius and the angle between the view direction and the SDF normals. 
As computing the Hessian matrix to obtain the analytical solution is computationally expensive, we approximate the curvature radius with a triangle from adjacent points. With the proposed bias-aware SDF to density transformation, our method reconstructs fine details faithfully for indoor scenes, as shown in Fig~\ref{fig:teaser}. 

In summary, we make the following contributions: 
\begin{itemize}
\item We identify the key reasons for previous methods failing to recover thin and detailed structures in indoor scenes. Consequently, we introduce \textbf{DebSDF}, which utilizes uncertainty modeling for filtering large error monocular priors, guiding the ray sampling, and applying smoothness regularization adaptively for reconstructing detailed indoor scenes.
\item We propose a novel bias-aware SDF to density transformation for volume rendering, enabling the reconstruction of thin and detailed structures. 
\item Extensive experiments on four challenging datasets, i.e., ScanNet~\cite{dai2017scannet}, ICL-NUIM~\cite{handa2014benchmark}, Replica~\cite{straub2019replica}, and Tanks and Temples~\cite{Knapitsch2017}, validate the effectiveness of our method quantitatively and qualitatively. 
\end{itemize}

The rest of the paper is organized as follows: In Sec. 2, we introduce some related works and contrast them with our neural indoor reconstruction methods to highlight our contribution. In Sec. 3, we introduce the details of our method, including uncertainty-guided prior localization, filtering, ray sampling, regularization, and our bias-aware SDF to density transformation. We conduct extensive experiments to validate their effectiveness on multiple challenging datasets in Sec. 4 and conclude our work in Sec. 5.

\section{Related works}\label{sec:related_works}
\noindent\textbf{Multi-view stereo.}
Multi-view stereo is a technique in computer vision that involves reconstructing a 3D scene from multiple 2D images captured from different viewpoints. Due to its wide-ranging applications in various fields, such as robotics, augmented reality, and computer graphics, it has been extensively studied within the field of computer vision. The classic multi-view stereo algorithms \cite{schonberger2016pixelwise,newcombe2011kinectfusion,merrell2007real,schonberger2016structure, bleyer2011patchmatch} employ patch matching across multiple images to estimate the depth \cite{schonberger2016pixelwise, bleyer2011patchmatch} of each pixel. Some methods \cite{paschalidou2018raynet,ulusoy2015towards,seitz1999photorealistic} utilize voxel representation to model shapes. However, these methods often encounter difficulties with the texture-less regions in the indoor scene since depth estimation and patch matching are difficult in these areas.

\begin{figure*}[h]
\centering
\includegraphics[width=0.95\textwidth, height=5.8cm]{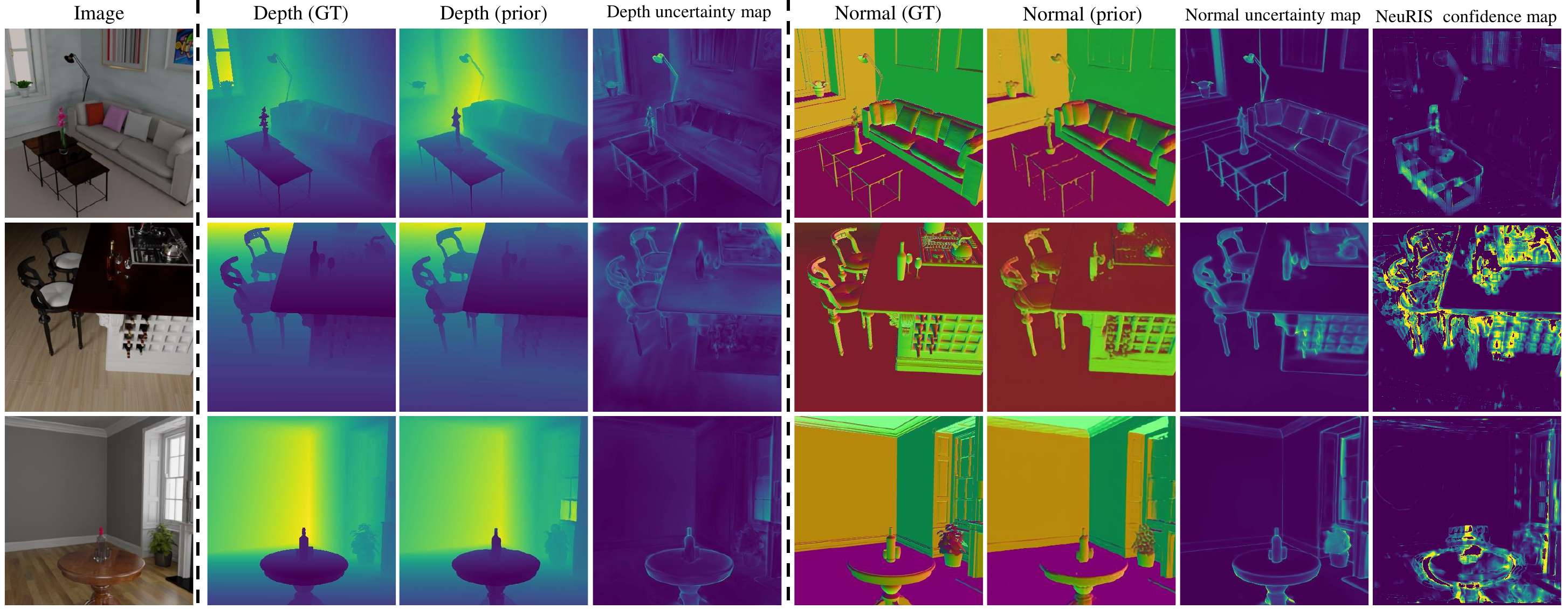}
\caption{The rendered uncertainty map can localize the regions where the monocular prior is inaccurate, which usually corresponds to thin structures with high levels of detail. Compared with NeuRIS \cite{wang2022neuris}, our model is capable of generating more reasonable results.
}
\label{fig:filter_mask}
\end{figure*}

Recent advancement of multi-view stereo focusing on the usage of deep learning approaches.
Some learning-based multi-view stereo methods replace some steps in the classic MVS pipeline. \cite{yu2020fast,yao2018mvsnet,huang2018deepmvs,yao2019recurrent} leverage a 3D CNN for depth estimation, yielding superior results. Most of these methods apply the MVSNet \cite{yao2018mvsnet} as the backbone, which is computationally expensive due to the large number of parameters involved in 3D CNN. Some works also apply the pyramid structure \cite{gu2020cascade,liao2021adaptive,cheng2020deep} to reduce the parameter quantity, while other works incorporate patch matching \cite{wang2021patchmatchnet}, attention mechanism \cite{ding2022transmvsnet}, and semantic segmentation \cite{xu2021self} for performance improvement. 
However, these approaches often suffer from geometry inconsistency since the depth maps are estimated individually for each view, resulting in noisy surfaces with many hollows.

\noindent\textbf{Neural scene representations.}
Instead of the multi-stage pipeline, some learning-based approaches propose an end-to-end framework to reconstruct 3D scenes based on the truncated signed distance function (TSDF) representation. Atlas \cite{murez2020atlas} extracts the image features by 2D CNN and back projects features from each view to the same 3D voxel volume. A 3D CNN is applied to infer the TSDF and labels of each point. NeuralRecon \cite{sun2021neuralrecon} utilizes a recurrent network for feature fusion from previous fragments sequentially. This can enhance the ability to both represent global and local details and reconstruct 3D scenes in real time. TransformerFusion \cite{bozic2021transformerfusion} applies the transformer for more efficient feature fusion from multi-view images, which benefits from the attention mechanism. 

Differing from treating the images as input to extract features by CNN, some methods \cite{yariv2021volume, wang2021neus, oechsle2021unisurf, mildenhall2021nerf, wei2021nerfingmvs, yariv2020multiview, barron2021mip, zhang2023towards, li2023neuralangelo} represent the scene as a coordinate-based implicit neural network and apply the differentiable neural rendering technique for supervision. Benefiting from the contiguous representation, these methods can generate arbitrary reconstruction results with only a simple fully connected deep network. Most of these methods have been integrated into open-source frameworks such as NeRFStudio~\cite{nerfstudio} and SDFStudio~\cite{Yu2022SDFStudio} to promote the development of neural fields.
Among them, NeRF maps the 5D coordinate to the volume density, and view-dependent emitted radiance, which achieves outstanding novel view synthesis performance, but the geometry is noisy. IDR \cite{yariv2020multiview} represents the geometry as the zero level set of a neural network and utilizes the surface rendering with the masks of each view to simultaneously optimize the geometry, view-dependent color, and camera poses. VolSDF \cite{yariv2021volume}, NeuS \cite{wang2021neus}, and HF-NeuS \cite{wang2022hf} apply SDF to represent the geometry and transform SDF to volumetric density such that the volume rendering can be used to render the image from each viewpoint. However, these methods usually fail in the reconstruction of the texture-less regions, such as the pure white wall in the indoor scene. This would lead the optimization to the local optimum, which indicates incorrect geometry results.

\noindent\textbf{Indoor scene reconstruction with priors.}
To tackle the aforementioned local minimization issue caused by the texture-less regions, some NeRF-based methods corporate prior information such as depth smoothness \cite{niemeyer2022regnerf}, semantic mask \cite{zhi2021place, jain2021putting}, and depth prior \cite{deng2022depth,roessle2022dense,wei2021nerfingmvs}. \cite{deng2022depth,roessle2022dense,wei2021nerfingmvs} integrate the depth information from the Structure-from-motion (Sfm) technique to regularize the geometry optimization. DSNeRF \cite{deng2022depth} and NerfingMVS \cite{wei2021nerfingmvs} apply the sparse depth map from COLMAP \cite{schonberger2016pixelwise} to regularize the optimization. DDPNeRF \cite{roessle2022dense} utilizes a pre-trained depth completion network to predict the dense depth map and a 2D uncertainty map to regularize the estimated depth and standard deviation of rendering weights. However, the predicted 2D uncertainty map from the pre-trained network suffers from the domain gap problem and cannot keep consistency in 3D space. The standard deviation of rendering weights is not suitable to apply to the SDF-based methods \cite{yariv2021volume,wang2021neus, oechsle2021unisurf} since the SDF-to-density transformation constrains the density distribution. 

Although the above NeRF-based methods can generate high-quality novel view synthesis results, the geometry reconstruction results are still noisy and incomplete. Some previous works \cite{coughlan1999manhattan,wang2022neuris,yu2022monosdf} focus on generating high-fidelity geometry by applying monocular priors such as depth and normal estimated by the pre-trained network. Manhattan-SDF \cite{guo2022neural} regularize the surface normal at the wall and floor regions under the Manhattan world assumption \cite{coughlan1999manhattan}. NeuRIS \cite{wang2022neuris} adopts the normal prior for regularization and uses an adaptive manner to impose the normal prior based on the assumption that the normal priors are not faithful at the pixels with rich visual features. MonoSDF \cite{yu2022monosdf} utilizes both the depth and normal priors predicted by the Omnidata model \cite{eftekhar2021omnidata} for 3D indoor scene reconstruction. The aforementioned previous works use the geometry prior for regularization and \cite{wang2022neuris} filter the normal prior according to the image feature. This can significantly improve the reconstruction quality in smooth and simple regions but still can not reconstruct the complex detailed surfaces well. The TUVR \cite{zhang2023towards} utilizes the MVS network \cite{yao2018mvsnet, gu2020cascade} for depth prior prediction and applies the cosine of the normal and ray direction to reduce the bias. The \cite{ye2023self} applies the self-supervised super-plane constraint by exploring the indoor reconstruction without the geometry cues. Similar to \cite{ye2023self}, the HelixSurf \cite{liang2023helixsurf} uses an unsupervised MVS approach to predict the depth point for intertwined regularization. The \cite{dong2023fast} directly uses the signed distance function in sparse voxel block grids for fast reconstruction without MLPs.

These methods \cite{yu2022monosdf, wang2022neuris, zhang2023towards} cannot robustly filter the inaccurate monocular priors, which is important to reconstruct the thin and detailed surfaces. Besides, other methods \cite{zhang2023towards, yariv2021volume, wang2021neus} ignore the bias in SDF-based rendering caused by the curvature of SDF. In this work, we focus on these problems to boost the 3D reconstruction of thin structures.

\section{Our Method}\label{sec:methods}
Previous works \cite{wang2022neuris, yu2022monosdf, guo2022neural} have shown that regularizing the optimization with geometry priors can significantly improve the reconstruction quality at texture-less areas such as the wall and floor within the implicit neural surface representation framework and SDF-based volume rendering. However, it is still difficult to reconstruct the complex and detailed surface, especially when it is less observed in the indoor scene, such as the legs of the chair. We analyze that there are some reasons for this problem: (i) The obtained geometry priors have significantly larger errors in these regions than in other planar regions. (ii) Areas of fine detail occupy a small area in the indoor scene, so the unbalanced sampling harms the reconstruction quality. (iii) Applying the smooth regularization indiscriminately degrades the reconstruction of high-frequency signals. (iv) The SDF-based volume rendering has geometry bias resulting from the curvature of SDF, which leads to the elimination of fine and detailed thin geometry structures, especially with regularization from monocular geometry prior. To tackle these problems, we propose \textbf{DebSDF}, which filters the inaccurate monocular priors and uses a bias-aware transformation from SDF to density to reduce the ambiguity of density representation such that the elimination of the fine structure problem can be solved.

\begin{figure*}[h]
\centering
\includegraphics[width=0.95\textwidth]{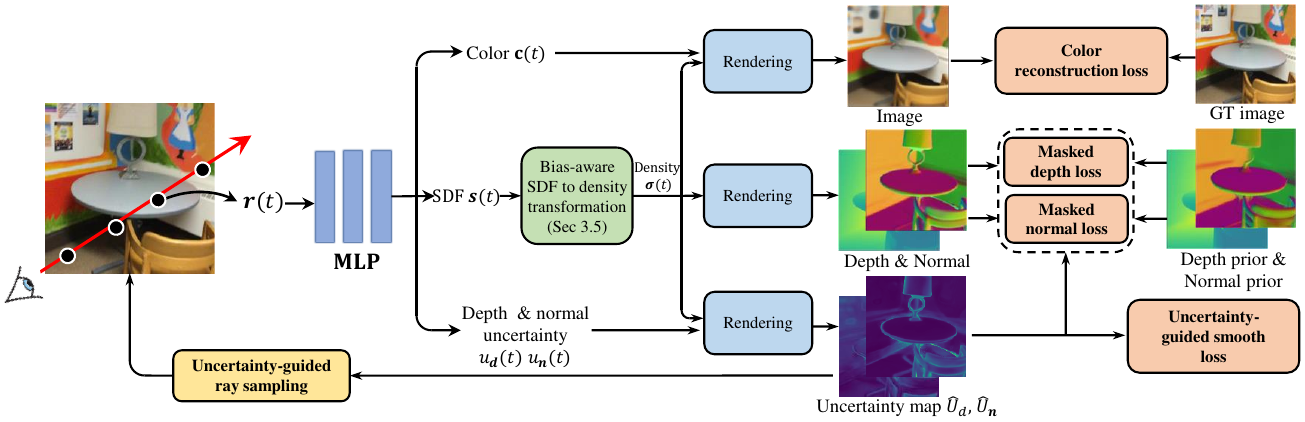}
\caption{The overview of our method. We propose the masked uncertainty learning to adaptively filter the geometry prior and localize the detailed and thin region in 5D space so that the small and thin structure would not be lost due to the wrong prior. Then, the localized uncertainty maps are utilized to guide ray sampling and smooth regularization to improve the reconstruction details of geometry. Besides, we analyze the bias in volume rendering caused by the transformation from SDF to density, which has a significant negative impact on the small and thin structure with geometry prior. A bias-aware SDF to density transformation is proposed to significantly reduce the bias for reconstructing small and thin objects.
}
\label{fig:overview}
\end{figure*}

\subsection{Preliminaries}
Following previous works \cite{yariv2021volume, yu2022monosdf}, we apply the implicit neural network to represent the geometry and radiance field and optimize this by differentiable volume rendering. Suppose a ray $\textbf{r}$ is cast from the camera location $\textbf{o}$ and passes through the pixel along the ray direction $\textbf{v}$. $N$ points are sampled on the ray and the $i$-th point is defined as $\textbf{r}(t_i)=\textbf{o} + t_i\textbf{v}$ where the $t_i$ is the distance to camera. The SDF $s_i$ and color value $\textbf{c}_i$ corresponding to the $\textbf{r}(t_i)$ is predicted by the implicit network. To apply the volume rendering technique, the transformation from SDF $s$ to density can be defined as the Laplace CDF\cite{yariv2021volume}:
\begin{equation}
    \sigma(s)=
    \begin{cases}
    \frac{1}{2\beta}\exp(-\frac{s}{\beta}) & \text{if}\  s > 0\\
    \frac{1}{\beta} - \frac{1}{2\beta}\exp(\frac{s}{\beta}) & \text{if} \ s \leq 0
    \end{cases}
\label{eq:sdf2density}
 \end{equation}
or the Logisitc CDF \cite{wang2022hf}, which proved with less bias than the former:
\begin{equation}
    \sigma(s) = \frac{1}{\beta} \cdot \frac{1}{1 + \exp(-\frac{s}{\beta})}
\end{equation}
where the $\beta$ is a learnable parameter.

The rendered color \cite{mildenhall2021nerf} of the ray $\textbf{r}$ in the space is:
\begin{equation}
    \hat{\textbf{C}}(\textbf{r}) = \sum_{i=1}^{N}T_i\alpha_i\textbf{c}_i,
\label{recon_loss}
\end{equation}
where $T_i$ and $\alpha_i$ denote the transparency and alpha value at the $i^{th}$ point on the ray $\textbf{r}$, respectively \cite{mildenhall2021nerf}:
\begin{equation}
     \quad T_i=\prod_{j=1}^{i-1}(1-\alpha_i), \quad \alpha_i = 1 - \exp(-\sigma_i\delta_i)
\label{eq:transmit}
\end{equation}
The rendered depth $\hat{D}(\textbf{r})$ and normal $\hat{N}(\textbf{r})$ corresponding to the surface intersecting the ray $\textbf{r}$ are \cite{yu2022monosdf, wei2021nerfingmvs, roessle2022dense}:
\begin{equation}
     \hat{D}(\textbf{r}) = \sum_{i=1}^{N}T_i\alpha_i t_i, \quad \hat{\textbf{N}}(\textbf{r}) = \sum_{i=1}^{N}T_i\alpha_i \textbf{n}_i
\label{eq:geo}
\end{equation}
where the $\textbf{n}_i$ is the SDF gradient of $i$-{th} point on ray $\textbf{r}$. 

The depth estimation network predicts the depth only up to scale, so the scale $w$ and the shift $q$ computed by the least-squares method \cite{eigen2014depth} is applied to normalize the depth prior, which is denoted as $D(\textbf{r})=wD'(\textbf{r})+q$. The $D'(\textbf{r})$ is the depth prior predicted by the pre-trained depth estimation network. The monocular depth and normal loss function for regularization of previous works \cite{yu2022monosdf, wang2022neuris} are:
\begin{equation}
\begin{aligned}
    & \mathcal{L}_{\textbf{depth}} = \sum_{\textbf{r}\in \mathcal{R}}\|w\hat{D}(\textbf{r})+q - D(\textbf{r})\|^2   \\
    \mathcal{L}_{\textbf{normal}} & = \sum_{\textbf{r}\in \mathcal{R}}\|\hat{\textbf{N}}(\textbf{r})- \textbf{N}(\textbf{r})\|_1 + \|1-\hat{\textbf{N}}(\textbf{r})^{T} \textbf{N}(\textbf{r})\|_1  \\
\end{aligned}
\end{equation}
where the $D(\textbf{r})$ and $\textbf{N}(\textbf{r})$ are the depth and normal prior obtained from the pre-trained Omnidata model \cite{eftekhar2021omnidata}.


\subsection{Uncertainty Guided Prior Filtering}
Since the monocular priors provided by the pre-trained models, such as Omnidata \cite{eftekhar2021omnidata} and SNU \cite{bae2021estimating}, are not perfectly accurate, it is necessary to apply an adaptive strategy to filter the monocular prior. NeuRIS \cite{wang2022neuris} filters the monocular prior based on the assumption that the regions where the monocular priors are not faithful typically consist of high-frequency features or irregular shapes with relatively rich visual features in the input images. However, this assumption does not generalize well since the monocular prior could be faithful at the simple planar regions, and some planar surfaces also have high-frequency appearance features, such as the wall with lots of texture details. Instead of applying the image feature for monocular prior filtering, we utilize the prior uncertainty from multi-view to filter the faithful prior. Specifically, the prior from a viewpoint is considered to be inaccurate if it has a large variance from other viewpoints. Besides, this also cannot guarantee the occlusion-aware property since whether a point on the ray is visible to other views is unknown. Based on the observation that the inaccurate priors usually have a large variance from multiple viewpoints, we introduce the masked uncertainty learning loss function to model this variance.

Following the \cite{yariv2021volume, yu2022monosdf}, the SDF values are predicted by an coordinate-based implicit network $f_g$:
\begin{equation}
    s, \textbf{z} = f_g(\textbf{x})
\end{equation} 
where the $\textbf{z}$ is the feature vector and the $\textbf{x}$ is the point coordinate on the ray $\textbf{r}$. 

To model the prior uncertainty of a pixel, a straightforward approach is to apply the variance of geometry prior from different viewpoints at the same point on the surface as the prior uncertainty. However, this approach models the uncertainty view-independent, which is the main drawback since only the viewpoints with faithful priors need to be filtered. Besides, it is still unknown whether the points on a queried ray are visible to the other views due to occlusion. The uncertainty computed by the pre-trained depth or normal estimation network \cite{roessle2022dense, bae2021estimating} has low accuracy because of the domain gap. Due to this reason, we model the uncertainty of the prior as a view-dependent representation by the volume rendering \cite{shen2022conditional, pan2022activenerf}.

Suppose the uncertainty scores at each point for modeling the variance corresponding to the monocular prior are $u_d \in \mathbb{R}$ and $\textbf{v}_n \in \mathbb{R}^3$, respectively. We apply the view-dependent color network to predict the uncertainty scores:
\begin{equation}
    \textbf{c}, u_d, \textbf{v}_n = f_c(\textbf{x}, \textbf{v}, \textbf{n}, \textbf{z})
\end{equation}

We compute the depth uncertainty score $\hat{U}_d(\textbf{r})$ and normal uncertainty vector
$\hat{\textbf{V}}_n(\textbf{r})$ of the pixel corresponds to the ray $\textbf{r}$ based on the volume rendering:
\begin{equation}
     \hat{U}_d(\textbf{r}) = \sum_{i=1}^{N}T_i\alpha_i u_{di}, \quad \hat{\textbf{V}}_n(\textbf{r}) = \sum_{i=1}^{N}T_i\alpha_i \textbf{u}_{ni} .
\label{eq:un}
\end{equation}
We compute the normal uncertainty score $\hat{U}_n(\textbf{r})$ as the mean value of the normal uncertainty vector $\hat{\textbf{V}}_n(\textbf{r})$ with 3 dimensions.

Moreover, we design the loss function based on the uncertainty learning perception with the mask to optimize the implicitly represented uncertainty field. The mask is applied to filter the negative impact from monocular prior with large uncertainty, which indicates a large probability of being inaccurate.


\noindent\textbf{Masked Depth loss.}
The $\hat{D}(\textbf{r})$ is the predicted geometry prior. 
We design the loss function as:
\begin{equation}
    \mathcal{L}_{\textbf{Mdepth}} = \ln|\hat{U}_d(\textbf{r})| + \frac{\Omega(\hat{U}_d(\textbf{r}), \tau_d) \bigodot |\hat{D}(\textbf{r})-D(\textbf{r})|}{|\hat{U}_d(\textbf{r})|}
\end{equation}
where $\Omega(U, \tau) \bigodot F$ is an adaptive gradient detach operation. The gradients from loss function would be detached if $U > \tau$ while the gradients are not detached if $U \leq \tau$. This indicates that the geometry prior is not utilized in the regions with high uncertainty while the uncertainty score will still be optimized.

\noindent\textbf{Masked Normal loss.}
Similar to the depth loss, the normal prior is transformed into the world coordinate space for regularization and computing the uncertainty.
\begin{equation}
    \mathcal{L}_{\textbf{Mnormal}} = \ln\hat{U}^2_n(\textbf{r}) + \frac{ \Omega(\hat{U}_n(\textbf{r}), \tau_n) \bigodot \|\hat{N}(\textbf{r})-N(\textbf{r})\|_2}{\hat{U}^2_n(\textbf{r})}
\end{equation}
Specifically, the inconsistency can be considered as the adaptive weight to adjust the monocular prior to each pixel. Low weight is applied to the prior with a large inconsistency.


\noindent\textbf{Color reconstruction loss.} Optimize the scene representation though the observation in 2D space.
\begin{equation}
    \mathcal{L}_{\textbf{rgb}} = \sum_{\textbf{r}\in \mathcal{R}}\|\hat{\textbf{C}}(\textbf{r})-\textbf{C}(\textbf{r})\|_1
\end{equation}

\noindent\textbf{Eikonal loss.} Following the \cite{gropp2020implicit}, the Eikonal loss is applied such that the property of SDF can be satisfied.
\begin{equation}
        \mathcal{L}_{\textbf{eik}} = \sum_{\textbf{x}\in\mathcal{X}}(\|\nabla f_g(\textbf{x})\|_2 - 1)^2 
\end{equation}
where the $\mathcal{X}$ is the set of points uniformly sampled in the 3D space and the regions near the surface.

The uncertainty map can estimate the multi-view consistency of depth and normal prior. If the most geometry prior to a region is inaccurate, 2 edge cases may occur: 
\begin{itemize}
\item The geometry prior from a viewpoint is inaccurate, but the corresponding uncertainty is low.
\item The geometry prior from a viewpoint is accurate, but the corresponding uncertainty is high.
\end{itemize}
Since the geometry prior is predicted by a pre-trained model and the real prior is unavailable, it is impossible to filter the inaccurate prior perfectly. However, our uncertainty-guided prior filtering can still filter the vast majority of incorrect prior. Since the inaccurate prior is unlikely to be multi-view consistent. After the negative influence of inaccurate prior is reduced, the color reconstruction loss would gradually dominate. Based on this analysis, these 2 edge cases would become increasingly rare along with the training.

\subsection{Uncertainty-Guided Ray Sampling}
Though applying the geometry prior with the uncertainty-based filter can improve the 3D reconstruction quality since inconsistency prior for some pixels can be filtered out, the fine and detailed structures are still hard to reconstruct. We observe that the thin object structures only occupy a small area in the image, so the probability of being sampled is low. In contrast, the texture-less and planar regions occupy most of the room's area and can already be reconstructed with high fidelity by a small number of ray samples, which benefits from the geometry prior supervision. Sampling more rays on these simple geometry surfaces than complex and fine geometry causes computational waste. 

Localizing the thin geometry surface, which usually corresponds to the high-frequency surface, is perceptual. A straightforward approach is to apply a high pass filter or keypoint extractor to localize the location of high-frequency signal for sampling, which generates incorrect results at the planar surfaces with high-frequency color appearance. Even with the auxiliary prior information such as the geometry prior \cite{eftekhar2021omnidata}, directly utilizing the high-frequency part from geometry prior is still inaccurate since the geometry prior would miss some detailed structure, which indicates predicting these regions as the smooth and planar surfaces. This can be observed in Fig. \ref{fig:teaser} where the chair legs are lost.



Based on this analysis, we compute the blend uncertainty score $A(\textbf{r})$ of each pixel by combining both inconsistency representations for depth and normal:
\begin{equation}
    A(\textbf{r}) = \hat{U}_d^{1-\lambda}(\textbf{r})\hat{U}_n^{{\lambda}}(\textbf{r})
\label{eq:importance}
\end{equation}
where the $\lambda$ is a hyper-parameter. The blend uncertainty score is utilized as the guidance for the ray sampling.

\begin{figure}[t]
\centering
\includegraphics[width=0.43\textwidth]{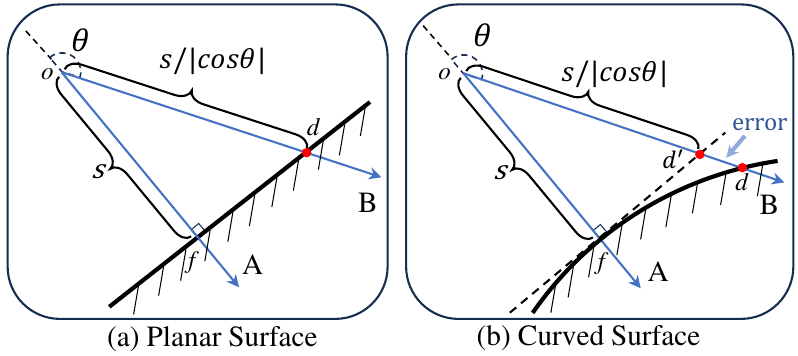}
\caption{
As shown in sub-figure (i), the TUVR \cite{zhang2023towards} applies the cosine between the ray direction and normal $\cos\theta$ to reduce the bias. Specifically, it transforms the SDF $s$ to the depth $s/|cos\theta|$ ($of\rightarrow od$) by assuming the ray intersects with the planar surface. However, TUVR \cite{zhang2023towards}, NeuS \cite{wang2021neus}, and HF-NeuS \cite{wang2022hf} only consider the planar surface, which ignores the curvature of the surface. As shown in sub-figure (ii), the $dd'$ indicates the error between the rendered depth and the real depth, which causes the biased volume rendering.
}
\label{fig:prob_formula_3.5}
\end{figure}

\begin{figure*}[h]
\centering
\includegraphics[width=0.88\textwidth]{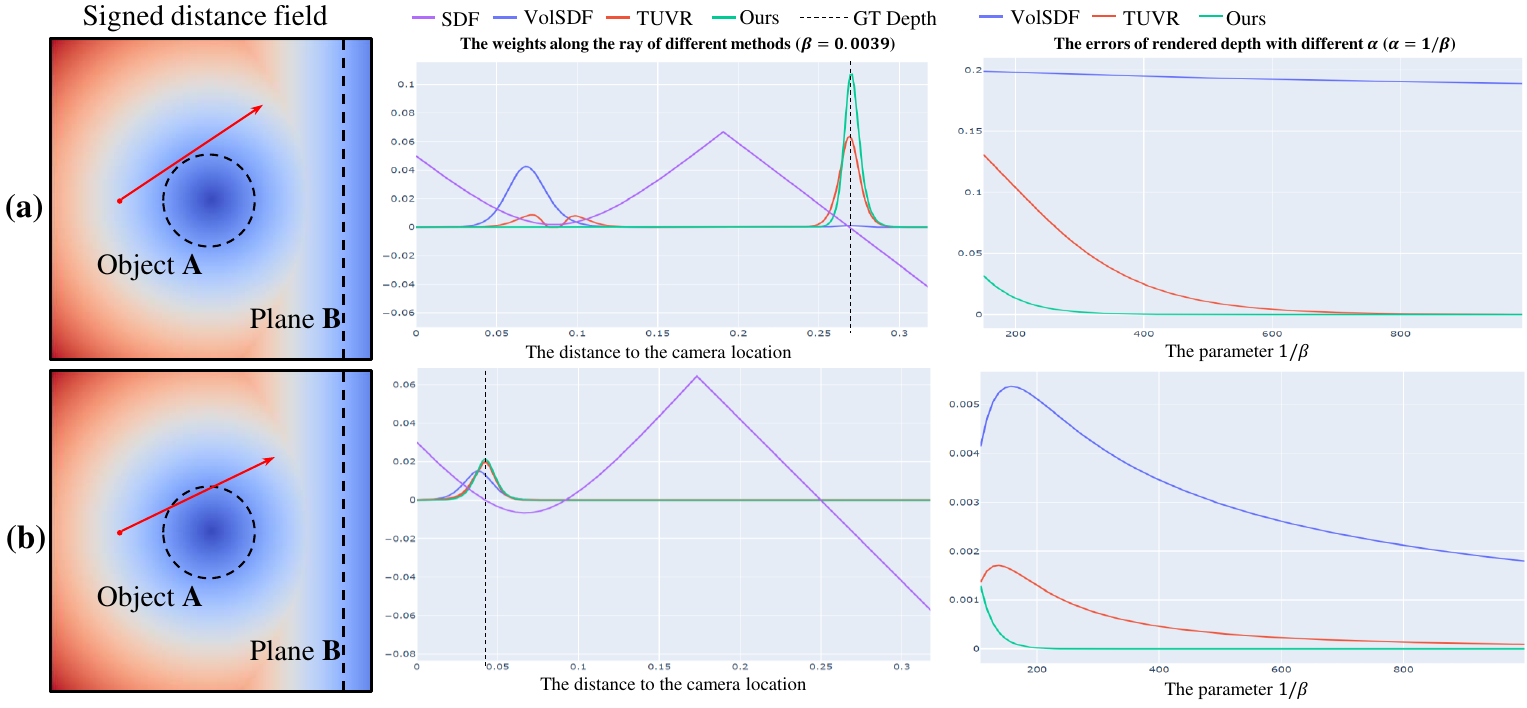}
\caption{We demonstrate 2 toy cases to simulate the SDF-based rendering by applying Logistic CDF when a ray brushes past and intersects with a small object. The weight function along the ray demonstrates that our method does not have a peak if there is no intersection with the surface. The sub-figures of the right column show the rendered depth corresponding to different $\alpha$ values. Our method can achieve smaller errors than previous works such as the TUVR \cite{zhang2023towards} and the VolSDF \cite{yariv2021volume}. We do not demonstrate NeuS \cite{wang2021neus} since the TUVR \cite{zhang2023towards} demonstrates that its bias is smaller than NeuS \cite{wang2021neus}. 
}
\label{fig:prob_formula}
\end{figure*}

Compared with inferring the confident maps for monocular prior filtering and ray sampling by only detecting the high-frequency image features, our method infers the uncertainty maps by using the information from multiple viewpoints implicitly. For each ray $\textbf{r}$, the probability to be sampled is calculated as:
\begin{equation}
    p(\textbf{r}_i) = \frac{A(\textbf{r}_i)}{\sum_i A(\textbf{r}_i)}
\end{equation}

\subsection{Uncertainty-Guided Smooth Regularization}
To avoid the reconstructed surfaces being too noisy, smooth regularization \cite{niemeyer2020differentiable} is widely applied to reduce the floater on the surfaces.
The smooth loss \cite{niemeyer2020differentiable} requires the gradients of SDF to be the same in a local region, which can reduce the floaters near the surface. 

However, this regularization would also damage the fine details, which indicates that not all the surfaces in the indoor scene need to be smooth. According to this analysis, we utilize the smooth regularization term in an adaptive manner, which can not only keep the simple surfaces smooth but also preserve the fine details of complex geometry.

For each sampled ray $\textbf{r}$, the $\mathcal{S(\textbf{r})}$ denotes the points sampled near the surface along this ray. The smoothness loss term is:
\begin{equation}
    \mathcal{L}_{\textbf{smooth}} = \mathbf{M}(A(\textbf{r}), \tau_s)\sum_{\textbf{x}\in\mathcal{S}(\textbf{r})}\|\nabla f_g(\textbf{x})-\nabla f_g(\textbf{x} + \epsilon)\|
\end{equation}
where the $\epsilon$ is a random offset sampled on a Gaussian distribution $\mathcal{N}(0, \xi )$ whose $\xi$ is a small variance. The $\mathbf{M}(A(\textbf{r}, \tau_s))$ is a mask function:
\begin{equation}
    \mathbf{M}(A(\textbf{r}, \tau_s))=
    \begin{cases}
    1 & \text{if}\  A(\textbf{r})\leq \tau_s\\
    0 & \text{if} \ A(\textbf{r}) > \tau_s
    \end{cases}
\label{eq: mask}
\end{equation}

Apart from the points sampled on the rays, a small batch of points is uniformly randomly sampled in the indoor space, and the aforementioned smoothness regularization is also applied. Since the probability of sampling near the surface is very low, the uncertainty-based adaptive manner is not used. Besides, we do not apply the adaptive manner on the Eikonal loss since the SDF around the small and thin structure still satisfies this property.

\subsection{Bias-aware SDF to Density Transformation}
Previous works \cite{yariv2021volume,wang2021neus,oechsle2021unisurf} utilize a transformation from SDF to volumetric density and volume rendering technique for modeling the connection between the 3D geometry and the rendered images. All these methods are designed to model a weight function that is required to be unbiased for the zero-level set of SDF. 

However, these methods, such as VolSDF \cite{yariv2021volume} and NeuS \cite{wang2021neus}, still suffer from biased volume rendering. The TUVR \cite{zhang2023towards} analyzes the problem of these methods and proposes to scale the SDF with the angle between the normal and the ray direction, but it still ignores the bias caused by the curvature of SDF. Based on this, we think that the curvature of SDF should be considered for unbiased volume rendering.
In this section, we analyze the weight function from volume rendering near the "small object" and design a bias-aware transformation from SDF to density to improve the reconstruction quality of the small and fine structures.

\begin{figure*}[t]
\centering
\includegraphics[width=0.95\textwidth]{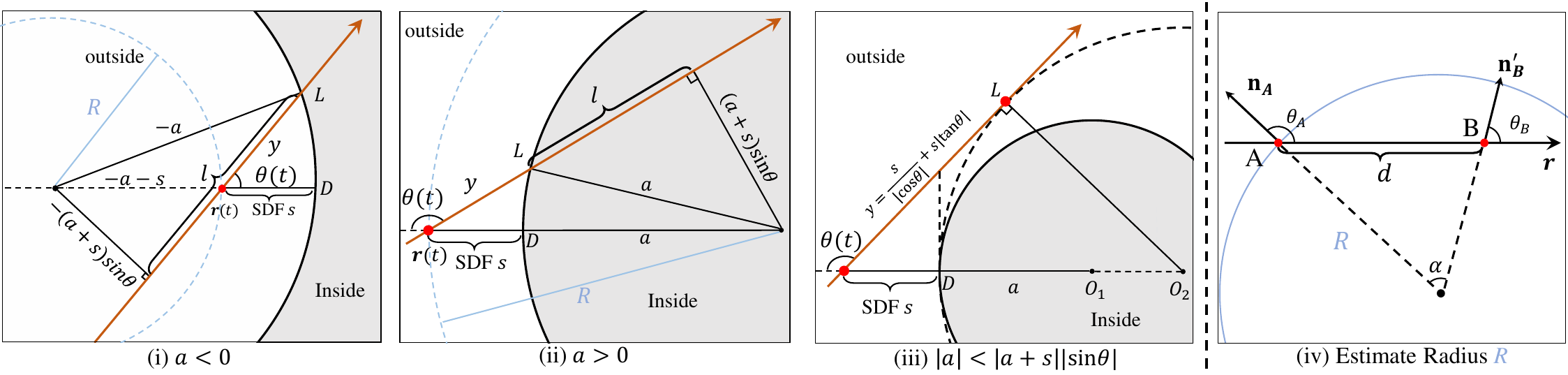}
\caption{The demonstration of our bias-aware transformation from SDF to density. The curvature radius $a>0$ indicates the intersection to the inwardly curved surface, while the curvature radius $a<0$ indicates the intersection with the outward curved surface. If the $|a|<|a+s||\sin\theta|$ is satisfied, this indicates that the ray and the surface nearby have no intersection point. The fourth sub-figure illustrates the curvature radius estimation by difference method. We apply 2 adjacent points on the ray to estimate the curvature radius of the previous point.
}
\label{fig:density_calib}
\end{figure*}

\subsubsection{The Analysis of Bias in the SDF to Density Transformation}
We can divide the design of ideal unbiased volume rendering into 2 steps. Firstly, design a transformation function from SDF to density which can ensure unbiased when the ray intersects with the surface vertically, which corresponds to the ray $A$ in Fig.  \ref{fig:prob_formula_3.5}. (a), the SDF value equal to the depth. This has been discussed in some previous works \cite{yariv2021volume, wang2021neus, wang2022hf, oechsle2021unisurf} while the Logistic transformation applied by HF-NeuS \cite{wang2022hf} can achieve unbiased transformation.
Secondly, since the ray may not intersect with the surface vertically, a transformation from the SDF to the depth is needed. The transformation proposed by TUVR \cite{zhang2023towards} is shown in Fig. \ref{fig:prob_formula_3.5}. It applies the cosine of the angle between the normal and the ray direction $\cos \theta$ for SDF transformation, which corresponds to ray $B$ in Fig. \ref{fig:prob_formula_3.5} (i). Besides, the NeuS \cite{wang2021neus} multiplies $\cos \theta$ by the interval between points on the ray to adjust this situation. However, these methods assume the intersection with a planar surface, which ignores the curvature of the surface. As shown in Fig. \ref{fig:prob_formula_3.5}, the bias still exists for the curved surface. Based on the aforementioned analysis, we propose a transformation from SDF to density, considering the curvature of SDF for reducing bias. Specifically, we assume that the surface is composed of multiple arcs and design a transformation from SDF to depth for the second step corresponding to the ($of\rightarrow od$) in Fig. \ref{fig:prob_formula_3.5} (ii).

We set 2 simple toy cases in Fig. \ref{fig:prob_formula} to demonstrate the benefits of considering the curvature. There is a small object $A$ and a plane $B$ in space. Case (a): A ray emits on the plane $B$ without intersecting with object A but being close to the object $A$. Case (b): A ray intersects with the object $A$. In sub-figure (a), though the ray does not intersect with object $A$, the weight functions along the ray of VolSDF \cite{yariv2021volume} and TUVR \cite{zhang2023towards} are still influenced by object $A$. Multiple peaks can be observed on the weight function when near object A. In sub-figure (b), it can be observed that the previous works have larger depth errors than our method. Especially the case shown in sub-figure (a) leads to multiple peaks, which have significant ambiguity for optimization. 

Because of this, the optimization of the masked depth loss, which minimizes the error between depth prior and rendered depth, would lead to the reduction of the density value around the peaks near the object $A$. And the masked normal loss suffers from the same problem as depth.
Though the points sampled around the multiple peaks along the ray are not inside the object $A$, it would still influence the SDF representation of object $A$ due to the local contiguity of the implicit neural network and the regularization of Eikonal loss.
This ambiguity causes the inefficient optimization of SDF for the fine and small objects in the indoor scene. Specifically, this problem causes the small and fine structures to be smaller or even disappear. 

Our model applies a better transformation from SDF to density, considering the curvature of the surface can have a smaller bias than other methods, which can have better reconstruction for the fine and detailed regions. 

\subsubsection{SDF to Density Mapping for Bias Reduction}

To tackle the aforementioned problem, we design an SDF to density transformation based on the analysis of the SDF curvature and the extreme point of weight function. Specifically, we consider a simple scenario: a ray $\textbf{r}$ intersects a circle of radius $a$ at point $L$. As shown in Fig. \ref{fig:density_calib}, the SDF of point $\textbf{r}(t)$ is $s$, but the ray reaches the circle surface through a distance of $y$. Our target is to design an SDF mapping function to replace the $s(t)$ in $\sigma(s(t))$ (Eq. \ref{eq:sdf2density}) as $y(t)$. We propose a function to map the SDF value $s$ to the distance $y$ in the transformation to density, which aims to reduce the negative influence from the curvature of SDF and the ray direction. This function can be divided into 2 situations: 
\begin{itemize}
    \item Ray intersects with the surface.
    \item Ray does not intersect the surface.
\end{itemize}

For the first situation, which is shown in Fig.\ref{fig:density_calib} (i, ii), the SDF mapping function is denoted as:
\begin{equation}
    y(t) = (a + s(t))|\cos\theta(t)| - \textbf{sign}(a)\cdot l
\label{eq:x}
\end{equation}
where the $\theta(t)$ is the angle between the view direction and the normal of SDF. The $a$ is the curvature radius of the point $D$ on the surface of the object $A$. Through Pythagorean theorem, the distance $l$ is denoted as:
\begin{equation}
    l = \sqrt{a^2-(a+s)^2\sin^2\theta}
\label{eq:l}
\end{equation}
Like the SDF, we define the distance function $y(t)$ as a negative number, so the above mathematical equations still hold when the $\textbf{r}(t)$ is inside the circle. This indicates that we assume the small local area on the surface can be approximated as a circular arc. For the planar surface, it can be considered the corresponding absolute value of the curvature radius is a very large number. The distance function $y(t)$ can be simplified to $y(t)=s(t)/|\cos\theta(t)|$, which is the same as TUVR \cite{zhang2023towards}. We apply the distance $y(t)$ as a kind of calibration to SDF $s(t)$ for volume rendering, so the density field of each point is computed as $\sigma(y(t))$.

Further, for the second situation which indicates the ray does not intersect with the surface, we design the mapping function based on the consideration that the ray tangent to another circle whose center is $O_2$ as shown in Fig.\ref{fig:density_calib} (iii). Specifically, if the number within the square root of Eq. \ref{eq:l} is less than 0, which corresponds to the situation $|a|< |a+s||\sin\theta|$, the ray does not intersect with the surface. For this situation, a naive solution is setting the $y(t)$ to infinity, which causes the discontinuity of the density field. To keep the density field contiguous, we design the distance $y(t)$ in this case as:
\begin{equation}
    y(t) = \frac{s(t)}{|\cos\theta(t)|} + s(t)|\tan\theta(t)|
\end{equation}
which indicates the tangency of another circle whose radius is larger than $a$.

Further, we apply the normal curvature radius of SDF at $\textbf{r(t)}$ to estimate the aforementioned curvature radius $a$. The curvature radius $a(t)$ corresponding to the point $\textbf{r}(t)$ is:
\begin{equation}
    a(t) = R(\textbf{r}(t), \textbf{v}) - s(t)
\end{equation}
where the $R(\textbf{r}(t), \textbf{v})$ \cite{wujun2007lines, novello2022exploring} is the normal curvature radius of SDF at point $\textbf{r}(t)$ toward the ray direction $\textbf{v}$.



\begin{figure*}[h!]
\centering
\includegraphics[width=0.99\textwidth, height=22cm]{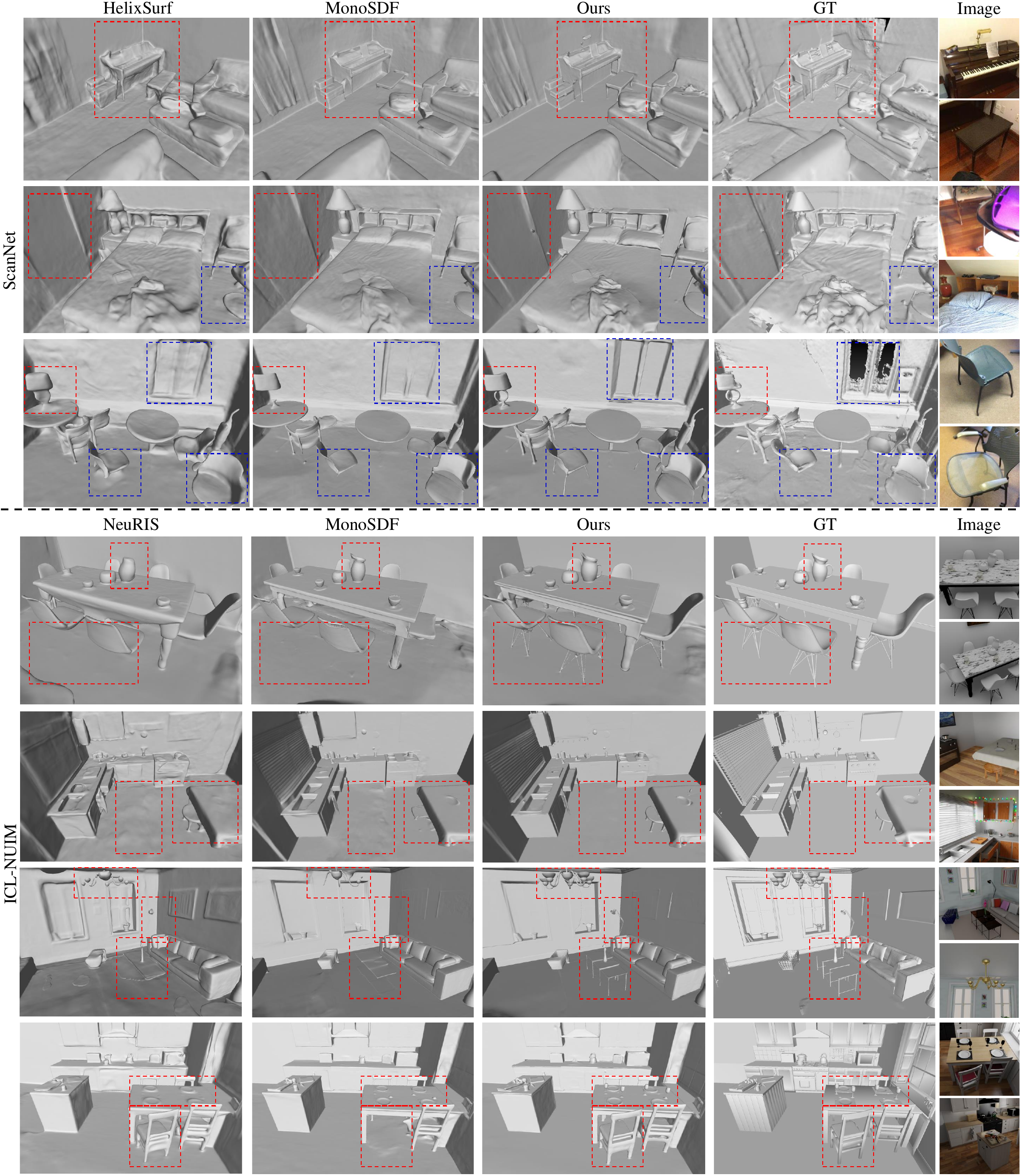}
\caption{The qualitative results on the ScanNet and ICL-NUIM dataset. We compare our method with previous state-of-the-art methods. Since the ground truth meshes of the ScanNet \cite{dai2017scannet} dataset are generated by the RGBD sensor, it is still not the complete and accurate reconstruction results, especially for the thin and detailed structures such as the leg and handrail of chairs and the lamp on the piano, which are shown in the blue box.
These can be confirmed by the column of images. 
We can observe that our method can reconstruct the thin and detailed surfaces significantly better than other methods.
}
\label{fig:comparison1}
\end{figure*}

\subsubsection{Curvature Radius Estimation}
Estimating the analytical solution of curvature radius requires computing the Hessian matrix, which is computationally unfriendly. We apply an approach to estimate the numerical solution of the curvature radius. As shown in the fourth sub-figure in Fig. \ref{fig:density_calib}, suppose there are 2 points $A$ and $B$ with distance $d$ on the ray $\textbf{r}$. The normals of point $A$ and $B$ are $\textbf{n}_A$ and $\textbf{n}_B$ respectively, and the $\textbf{n'}_B$ is the projection of $\textbf{n}_B$ on the plane determined by normal $\textbf{n}_A$ and ray direction $\textbf{v}$. The $\alpha$ is the angle between $\textbf{n}_A$ and $\textbf{n'}_B$.

The curvature radius $R$ of point $A$ can be estimated based on the Law of Sines:
\begin{equation}
    R = \chi(\textbf{n}_A, \textbf{n'}_B) \cdot d\cdot\frac{\sin\theta_B}{\sin\alpha}
\end{equation}
where the $\chi(\textbf{n}_A, \textbf{n'}_B)$ is the indicator function:
\begin{equation}
    \chi(\textbf{n}_A, \textbf{n'}_B) = 
    \begin{cases}
    1 & \text{if}\  \cos(\theta_A)\leq \cos(\theta_B)\\
    -1 & \text{if} \ \cos(\theta_A) > \cos(\theta_B)
    \end{cases}
\end{equation}

\subsubsection{Progressive Warm-up}
We design a progressive warm-up strategy to stabilize the training phase since the numerical estimation of the curvature radius is not stable at the beginning of training.
Specifically, we replace the $\cos\theta(t)$ to $\cos^{p(t)}\theta(t)$, where the $p(t)$ is a parameter progressive growing with training iterations from 0 to 1. Then correspondingly, the $\sin\theta$ is adjusted according to $\sin^2\theta+\cos^2\theta=1$. Moreover, since the sampling probability of the texture-less region is low, the growing $p(t)$ is designed as the product of a progressively increasing number of $p'$ and the point-wise uncertainty score, such that the $p$ of texture-less and planar regions increase slower than the detailed and important regions. The $p(t)$ corresponds to the point $\textbf{r}(t)$ is denoted as:
\begin{equation}
    p(t) = \min (p'\cdot u_d^{1-\lambda}(t) \cdot u_n^\lambda(t), 1)
\label{eq: n}
\end{equation}
where the $u_n(t)$ is the mean of the normal uncertainty score vector $\textbf{v}_n(t)$ corresponds to each point instead of the $\hat{\textbf{V}}_n$. 

\noindent\textbf{Optimization}
The total loss function to optimize the neural implicit geometry and color appearance network is:
\begin{equation}
    \mathcal{L} = \mathcal{L}_{\textbf{rgb}} + \lambda_1\mathcal{L}_{\textbf{eik}} + \lambda_2\mathcal{L}_{\textbf{smooth}} + \lambda_3\mathcal{L}_{\textbf{Mdepth}} + \lambda_4\mathcal{L}_{\textbf{Mnormal}}  .
\end{equation}
where the $\lambda_i$ is the hyper-parameters for weighting each term of the loss function. We set $\lambda_1=0.05$, $\lambda_2=0.005$, $\lambda_3=0.006$, and $\lambda_4=0.0025$.

\begin{figure*}[h]
\centering
\includegraphics[width=0.95\textwidth, height=11cm]{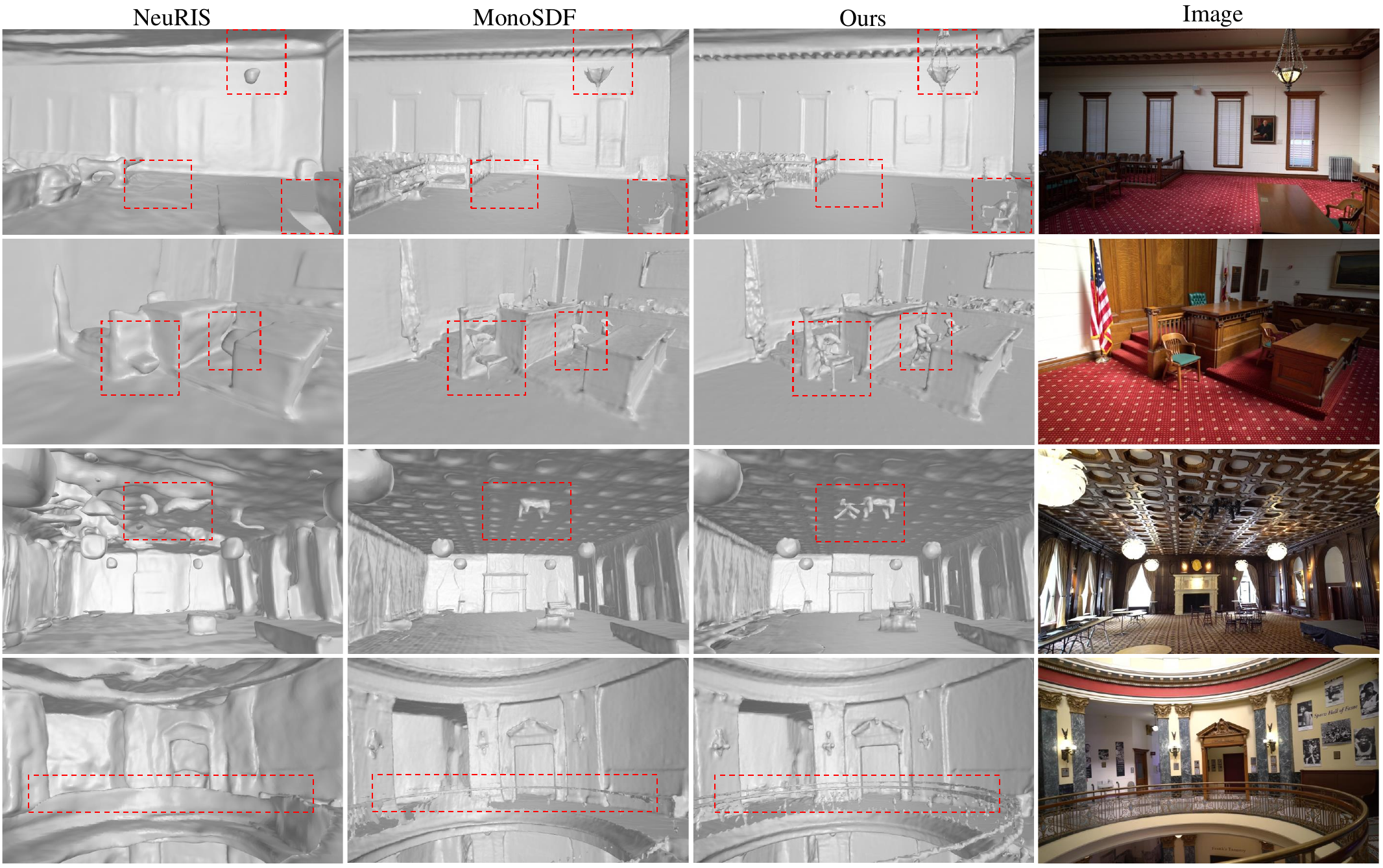}
\caption{We demonstrate the qualitative results on the Tanks and Temple \cite{Knapitsch2017} dataset. Our method reconstructs small objects better than MonoSDF \cite{yu2022monosdf}, such as the pendant, chairs, and handrails of stairs. As shown in the first column, our method can also adapt to the inaccurate prior on the floor, so there is not a large pit on the floor. The ground truth meshes of the Tanks and Temples \cite{Knapitsch2017} dataset are not released to the public.
}
\label{fig:comparison_tnt}
\end{figure*}

\section{Experiments}
\noindent\textbf{Datasets.}
We evaluate our method on the following 5 datasets: ScanNet \cite{dai2017scannet}, ICL-NUIM \cite{handa2014benchmark}, Replica \cite{straub2019replica}, Tanks and Temples \cite{Knapitsch2017} and DTU dataset\cite{aanaes2016large}. Among them, ScanNet, Replica, and Tanks and Temples are real-world indoor scene datasets, and the Tanks and Temples dataset has primary large-scale indoor scenes. We select 4 scenes from ScanNet for performance evaluation by following the setting of Manhattan-SDF \cite{guo2022neural}. The ICL-NUIM \cite{handa2014benchmark} is a synthetic indoor scene dataset with the ground truth mesh.

\noindent\textbf{Baselines.}
We compare our method with the following methods: (i) Neural volume rendering methods with prior, including MonoSDF \cite{yu2022monosdf}, NeuRIS \cite{wang2022neuris}, Manhattan-SDF \cite{guo2022neural}; (ii) Neural volume rendering methods without prior, including VolSDF \cite{yariv2021volume}, NeuS \cite{wang2021neus}, and Unisurf \cite{oechsle2021unisurf}; and (iii) Classical MVS reconstruction method: COLMAP \cite{schonberger2016pixelwise}.

\noindent\textbf{Metrics.}
By following the evaluation protocol of \cite{yu2022monosdf, wang2022neuris}, we use the following evaluation metrics: Chamfer Distance ($L_2$), F-score with 5cm threshold, and Normal Consistency. 

\begin{table}[t]
\setlength{\tabcolsep}{3.6pt} 
\caption{The quantitative results on the ScanNet \cite{dai2017scannet} dataset. Our method achieves state-of-the-art performance.}
\centering
\begin{tabular}{c|cccccc}
\toprule
Method  & Acc$\downarrow$            & Comp$\downarrow$           & Pre$\uparrow$           & Recall$\uparrow$        & Chamfer$\downarrow$        & F-score$\uparrow$       \\ \hline
COLMAP\cite{schonberger2016pixelwise}  & 0.047          & 0.235          & 71.1          & 44.1          & 0.141          & 53.7          \\
Unisurf\cite{oechsle2021unisurf} & 0.554          & 0.164          & 21.2          & 36.2          & 0.359          & 26.7          \\
VolSDF\cite{yariv2021volume}  & 0.414          & 0.120          & 32.1          & 39.4          & 0.267          & 34.6          \\
NeuS\cite{wang2021neus}    & 0.179          & 0.208          & 31.3          & 27.5          & 0.194          & 29.1          \\
M-SDF\cite{guo2022neural}   & 0.072          & 0.068          & 62.1          & 56.8          & 0.070          & 60.2          \\
NeuRIS\cite{wang2022neuris}  & 0.050          & 0.049          & 71.7          & 66.9          & 0.050          & 69.2          \\
MonoSDF\cite{yu2022monosdf} & \textbf{0.035} & 0.048          & 79.9          & 68.1          & 0.042          & 73.3          \\
TUVR\cite{zhang2023towards}    & 0.038          & 0.042          & 76.6          & 72.9          & 0.040          & 74.7          \\
HelixSurf\cite{liang2023helixsurf}    & 0.038          & 0.044          & 78.6          & 72.7          & 0.042          & 75.5          \\
Ours    & 0.036          & \textbf{0.040} & \textbf{80.7} & \textbf{76.5} & \textbf{0.038} & \textbf{78.5}  \\
\bottomrule
\end{tabular}
\label{table:scannet}
\end{table}

\begin{table}[]
\setlength{\tabcolsep}{4pt} 
\caption{The quantitative results on the ICL-NUIM \cite{handa2014benchmark} and Replica \cite{straub2019replica} dataset. Our method achieves state-of-the-art performance on both datasets. It is noted that the official MonoSDF \cite{yu2022monosdf} implementation on the Replica dataset utilizes manually selected masks to disable the monocular priors of some images due to serious inaccuracy. In this paper, to validate the ability of our model to filter the inaccurate prior automatically, we do not apply this operation to MonoSDF \cite{yu2022monosdf} and our method for fair comparisons.}
\centering
\resizebox{\linewidth}{!}{
\begin{tabular}{cc|ccc}
\toprule
\multicolumn{2}{l|}{}                                    & Normal C$\uparrow$       & Chamfer$\downarrow$        & F-score$\uparrow$        \\ \hline
\multicolumn{1}{c|}{\multirow{3}{*}{Replica}}  & Unisurf\cite{oechsle2021unisurf} & 90.96          & 0.049          & 78.99          \\
\multicolumn{1}{c|}{}                          & MonoSDF\cite{yu2022monosdf} & 92.69          & 0.045          & 81.46          \\
\multicolumn{1}{c|}{}                          & Ours    & \textbf{93.23} & \textbf{0.029} & \textbf{88.36} \\ \hline
\multicolumn{1}{c|}{\multirow{4}{*}{ICL-NUIM}} & NeuRIS\cite{wang2022neuris}  & 85.54          & 0.106          & 67.31          \\
\multicolumn{1}{c|}{}                          & MonoSDF\cite{yu2022monosdf} & 87.51          & 0.104          & 67.39          \\
\multicolumn{1}{c|}{}                          & TUVR\cite{zhang2023towards}    & 87.61          & 0.101          & 68.75          \\
\multicolumn{1}{c|}{}                          & Ours    & \textbf{88.34} & \textbf{0.092} & \textbf{77.38}       \\
\bottomrule
\end{tabular}
}
\label{table:icl_and_replica}
\end{table}

\begin{figure*}[h]
\centering
\includegraphics[width=0.95\textwidth, height=5.6cm]{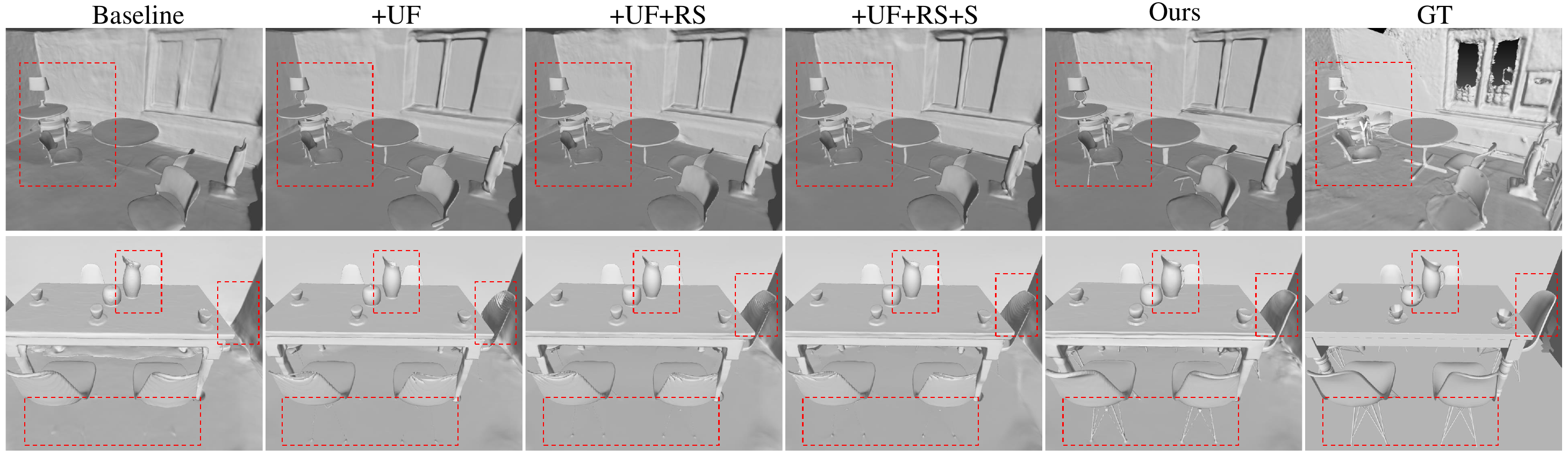}
\caption{The ablation studies of our method. "UF" denotes the uncertainty-guided prior filtering with the masked depth and normal loss; "RS" denotes the uncertainty-guided ray sampling; "S" denotes the uncertainty-guided smooth. Compared with the baseline, the reconstruction of the thin and detailed structure benefits from our proposed modules.
}
\label{fig:ablation}
\end{figure*}

\subsection{Implementation Details}
By following the MonoSDF \cite{yu2022monosdf}, we apply an MLP with 8 hidden layers as the geometry network to predict SDF and an MLP with 2 layers as the color network to predict the color field. Each layer has 256 hidden nodes. The Adam optimizer with a beginning learning rate of $5\times10^{-4}$ is utilized to optimize the network. The learning rate exponentially decays in each iteration. The size of the input image is $384\times384$, and the pre-trained Omnidata model \cite{eftekhar2021omnidata} is applied to generate the geometry prior. We use PyTorch \cite{paszke2019pytorch} to implement our model and train our model on one NVIDIA GeForce RTX 2080Ti GPU. We train our model for 200,000 iterations with 1024 rays sampled in each iteration. We set the $\lambda=0.9$ to balance the uncertainty map localized from the depth and normal prior. We do not apply the estimated uncertainty regions to guide ray sampling and smoothing for the first 40,000 iterations since the uncertainty localization is not stable at the initial stage. To filter out the wrong prior, we set $\tau_d=0.25$, $\tau_n=0.4$, and $\tau_s=0.3$. 

\begin{table}[t]
\caption{The quantitative results on Tanks and Temples dataset. * indicates using the hash encoding feature grids. The evaluation metric is the F-score, which is computed by the official website of the Tanks and Temple dataset.}
\centering
\resizebox{\linewidth}{!}{
\begin{tabular}{c|cccc|c}
\toprule
Method   & Auditorium    & Ballroom      & Courtroom      & Museum        & Mean          \\ \hline
NeuRIS\cite{wang2022neuris}   & 0.79          & 0.84          & 2.78           & 1.34          & 1.13          \\
MonoSDF\cite{yu2022monosdf}  & 3.09          & 2.47          & 10.00          & 5.10          & 5.16         \\
Ours     & \textbf{5.41} & \textbf{6.09} & \textbf{18.53} & \textbf{7.11} & \textbf{9.43} \\ \hline
MonoSDF*\cite{yu2022monosdf} & 3.17          & 3.70          & 13.75          & 5.68          & 6.58          \\
Ours*    & \textbf{7.45} & \textbf{4.21} & \textbf{19.33} & \textbf{7.64} & \textbf{9.66}      \\
\bottomrule
\end{tabular}
}
\label{table:tnt}
\end{table}

\begin{table*}[h!]
\caption{The ablation studies conducted on the ScanNet \cite{dai2017scannet} and ICL-NUIM \cite{handa2014benchmark} datasets. 
}
\centering
\resizebox{\linewidth}{!}{
\begin{tabular}{cccc|ccc|ccc}
\toprule
\multirow{2}{*}{\begin{tabular}[c]{@{}c@{}}Prior\\ Filtering\end{tabular}} & \multirow{2}{*}{\begin{tabular}[c]{@{}c@{}}Uncertainty-Guided\\ Ray Sampling\end{tabular}} & \multirow{2}{*}{\begin{tabular}[c]{@{}c@{}}Uncertainty-\\ Guided Smooth\end{tabular}} & \multirow{2}{*}{\begin{tabular}[c]{@{}c@{}}Bias-Aware SDF to \\ Density Transformation\end{tabular}} & \multicolumn{3}{c|}{ScanNet}                      & \multicolumn{3}{c}{ICL-NUIM}                      \\ \cline{5-10} 
                                                                                   &                                                                                   &                                                                                     &                                                                                              & Normal C$\uparrow$       & Chamfer$\downarrow$      & F-score$\uparrow$        & Normal C$\uparrow$      & Chamfer$\downarrow$      & F-score$\uparrow$        \\ \hline
                                                                                   &                                                                                   &                                                                                     &                                                                                              & 87.85          & 0.0429          & 73.30          & 87.51          & 0.1034          & 67.39          \\
\checkmark                                                                          &                                                                                   &                                                                                     &                                                                                              & 89.57          & 0.0402          & 76.82          & 88.12          & 0.0981          & 73.97          \\
\checkmark                                                                          & \checkmark                                                                         &                                                                                     &                                                                                              & 89.60          & 0.0399          & 77.12          & 87.99          & 0.0978          & 76.14          \\
\checkmark                                                                          & \checkmark                                                                         & \checkmark                                                                           &                                                                                              & 89.98          & 0.0399          & 77.30          & 88.30          & 0.0953          & 76.75          \\
\checkmark                                                                          & \checkmark                                                                         & \checkmark                                                                           & \checkmark                                                                                    & \textbf{90.21} & \textbf{0.0382} & \textbf{78.54} & \textbf{88.34} & \textbf{0.0928} & \textbf{77.59}     \\
\bottomrule
\end{tabular}
}
\label{table:ablations}
\end{table*}

\subsection{Performance Comparison with Other Baselines}
Regarding the baselines, COLMAP \cite{schonberger2016pixelwise} reconstructs the mesh from the depth point cloud by the Poisson Surface reconstruction algorithm \cite{kazhdan2013screened}. NeRF \cite{mildenhall2021nerf}, Unisurf\cite{oechsle2021unisurf}, NeuS \cite{wang2021neus}, and VolSDF \cite{yariv2021volume} apply the neural volume rendering technique for 3D reconstruction. The Manhattan-SDF \cite{guo2022neural}, NeurRIS \cite{wang2022neuris}, and MonoSDF \cite{yu2022monosdf} utilize the auxiliary data from pre-trained models. As shown in Table \ref{table:scannet} and Table \ref{table:icl_and_replica}, our method achieves state-of-the-art performance on multiple datasets. We also visualize the results of NeuRIS and MonoSDF, which are two state-of-the-art methods for indoor 3D reconstruction. The quantitative results are shown in Table \ref{table:scannet} and Table \ref{table:icl_and_replica}; our method achieves significant improvement compared with other methods, which indicates the importance of the uncertainty-guided prior filtering and bias-aware SDF to density transformation. As shown in Fig. \ref{fig:comparison1}, our method well reconstructs the small and thin structures, such as the leg and the handrail of chairs, the bracket of the lamp, and the cups on the Table. Further, the reconstruction of the texture-less surfaces is also improved for our method. As shown in the first and second rows of Fig. \ref{fig:comparison1}, the reconstructed floor and wall of our method are smoother than other methods. 

We also apply our methods on the Tanks and Temples dataset with/without hash encoding feature grids \cite{muller2022instant, yu2022monosdf}. As shown in Table \ref{table:tnt}, our method can achieve large improvement for both settings. The qualitative results are shown in Fig. \ref{fig:comparison_tnt}. It can be observed that our method prevents reconstruction results from being corrupted by the wrong geometry priors and simultaneously improves the performance at thin and detailed regions.

\begin{table*}[]
\caption{
We divide each image in the ICL-NUIM dataset \cite{handa2014benchmark} into the masked region and unmasked region with the mask from the blend uncertainty map and compute the metrics for depth and normal map evaluation. The masked regions approximately correspond to the thin and detailed surface, while the unmasked regions correspond to the simple planar surface. "Mesh" indicates extracting the normal and depth map from the reconstructed mesh, while "Volume Rendering" indicates extracting the normal and depth map from volume rendering of the SDF.}
\centering
\resizebox{\linewidth}{!}{
\begin{tabular}{c|c|cccccc|cccccc}
\toprule
\multicolumn{1}{l|}{\multirow{3}{*}{$\tau_s$}} & \multirow{3}{*}{Method} & \multicolumn{6}{c|}{Mesh}                                                                                                                                                                                                                                                                                                                                          & \multicolumn{6}{c}{Volume Rendering}                                                                                                                                                                                                                                                                                                                               \\ \cline{3-14} 
\multicolumn{1}{l|}{}                     & \multicolumn{1}{l|}{}                        & \multicolumn{3}{c|}{Masked region}                                                                                                                                                         & \multicolumn{3}{c|}{Unmasked region}                                                                                                                                  & \multicolumn{3}{c|}{Masked region}                                                                                                                                                         & \multicolumn{3}{c}{Unmasked region}                                                                                                                                   \\ \cline{3-14} 
\multicolumn{1}{l|}{}                     & \multicolumn{1}{l|}{}                        & \begin{tabular}[c]{@{}c@{}}depth\\ abs\_rel$\downarrow$\end{tabular} & \begin{tabular}[c]{@{}c@{}}normal\\ cos$\uparrow$\end{tabular} & \multicolumn{1}{c|}{\begin{tabular}[c]{@{}c@{}}normal\\ $L_1\downarrow$\end{tabular}} & \begin{tabular}[c]{@{}c@{}}depth\\ abs\_rel$\downarrow$\end{tabular} & \begin{tabular}[c]{@{}c@{}}normal\\ cos$\uparrow$\end{tabular} & \begin{tabular}[c]{@{}c@{}}normal\\ $L_1\downarrow$\end{tabular} & \begin{tabular}[c]{@{}c@{}}depth\\ abs\_rel$\downarrow$\end{tabular} & \begin{tabular}[c]{@{}c@{}}normal\\ cos$\uparrow$\end{tabular} & \multicolumn{1}{c|}{\begin{tabular}[c]{@{}c@{}}normal\\ $L_1\downarrow$\end{tabular}} & \begin{tabular}[c]{@{}c@{}}depth\\ abs\_rel$\downarrow$\end{tabular} & \begin{tabular}[c]{@{}c@{}}normal\\ cos$\uparrow$\end{tabular} & \begin{tabular}[c]{@{}c@{}}normal\\ $L_1\downarrow$\end{tabular} \\ \hline
\multirow{4}{*}{0.3}                      & NeuRIS \cite{wang2022neuris}                                       & 0.160                                                    & 0.637                                                & \multicolumn{1}{c|}{0.339}                                               & 0.030                                                    & 0.940                                                & 0.083                                               & 0.199                                                    & 0.662                                                & \multicolumn{1}{c|}{0.336}                                               & 0.057                                                    & 0.943                                                & 0.085                                               \\
                                          & MonoSDF \cite{yu2022monosdf}                                      & 0.167                                                    & 0.664                                                & \multicolumn{1}{c|}{0.323}                                               & 0.026                                                    & 0.962                                                & 0.054                                               & 0.180                                                    & 0.650                                                & \multicolumn{1}{c|}{0.331}                                               & 0.035                                                    & 0.959                                                & 0.057                                               \\
                                          & No Bias-aware                                & 0.136                                                    & 0.691                                                & \multicolumn{1}{c|}{0.291}                                               & 0.017                                                    & 0.965                                                & \textbf{0.049}                                      & 0.155                                                    & 0.693                                                & \multicolumn{1}{c|}{0.295}                                               & 0.03                                                     & 0.963                                                & 0.053                                               \\
                                          & Ours                                         & \textbf{0.117}                                           & \textbf{0.702}                                       & \multicolumn{1}{c|}{\textbf{0.287}}                                      & \textbf{0.016}                                           & \textbf{0.967}                                       & \textbf{0.049}                                      & \textbf{0.133}                                           & \textbf{0.700}                                       & \multicolumn{1}{c|}{\textbf{0.289}}                                      & \textbf{0.027}                                           & \textbf{0.964}                                       & \textbf{0.052}                                      \\ \hline
\multirow{4}{*}{0.4}                      & NeuRIS \cite{wang2022neuris}                                      & 0.185                                                    & 0.623                                                & \multicolumn{1}{c|}{0.345}                                               & 0.032                                                    & 0.934                                                & 0.089                                               & 0.243                                                    & 0.646                                                & \multicolumn{1}{c|}{0.344}                                               & 0.059                                                    & 0.938                                                & 0.089                                               \\
                                          & MonoSDF \cite{yu2022monosdf}                                      & 0.189                                                    & 0.643                                                & \multicolumn{1}{c|}{0.335}                                               & 0.029                                                    & 0.956                                                & 0.060                                               & 0.224                                                    & 0.622                                                & \multicolumn{1}{c|}{0.348}                                               & 0.037                                                    & 0.954                                                & 0.061                                               \\
                                          & No Bias-aware                                & 0.166                                                    & 0.674                                                & \multicolumn{1}{c|}{0.298}                                               & 0.019                                                    & 0.959                                                & 0.055                                               & 0.199                                                    & 0.660                                                & \multicolumn{1}{c|}{0.312}                                               & 0.031                                                    & 0.959                                                & 0.056                                               \\
                                          & Ours                                         & \textbf{0.142}                                           & \textbf{0.676}                                       & \multicolumn{1}{c|}{\textbf{0.300}}                                      & \textbf{0.018}                                           & \textbf{0.962}                                       & \textbf{0.054}                                      & \textbf{0.176}                                           & \textbf{0.661}                                       & \multicolumn{1}{c|}{\textbf{0.308}}                                      & \textbf{0.028}                                           & \textbf{0.960}                                       & \textbf{0.055} \\
\bottomrule
\end{tabular}   
}
\label{table: mask_eval}
\end{table*}

\subsection{Performance Evaluation of Different Regions}
\begin{figure}[]
\centering
\includegraphics[width=0.43\textwidth]{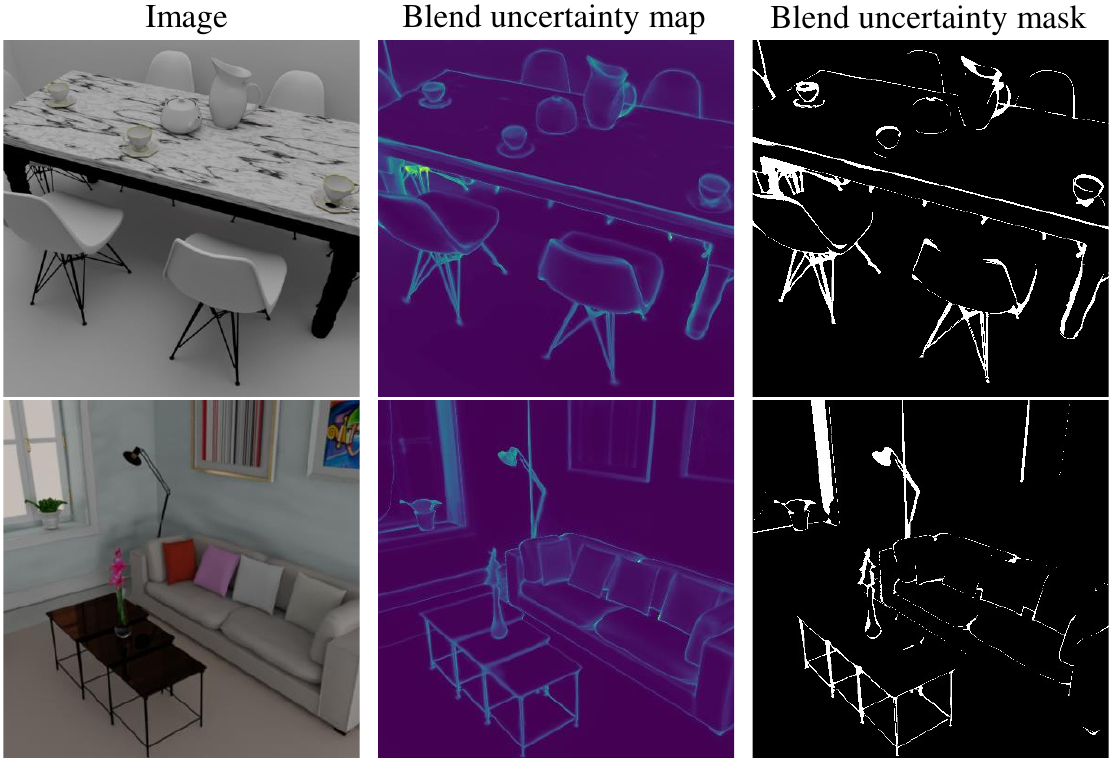}
\caption{The visualization of blend uncertainty map and mask. The mask can localize the fine and detailed regions in the indoor scene which would facilitate the ray sampling and smooth regularization.
}
\label{fig: blendmask}
\end{figure}

The aforementioned blend uncertainty map can be utilized to localize the thin and detailed structures in the image, so we divide each image into the masked and unmasked regions by setting different threshold $\tau_s$'s for the blend uncertainty mask obtained from Eq. \ref{eq: mask}, and the masked regions and unmasked regions correspond to the detailed surface and simple planar surface in the indoor scene, respectively. The visualizations of both parts are shown in Fig. \ref{fig: blendmask}. Then, we use the depth abs rel, normal cos similarity, and normal $L_1$ similarity metrics to evaluate the improvements of different regions for each viewpoint. we generate the predicted depth and normal map from reconstructed meshes obtained from Marching Cubes \cite{lorensen1987marching} and volume rendering. 

As shown in Table \ref{table: mask_eval}, the reconstruction quality improvements of our method on the complex detailed surfaces are more significant than those on the simple planar surfaces. For the bias-aware SDF to density transformation, the improvements of the masked regions are also more significant than the unmasked regions. This further validates the effectiveness of our method for the reconstruction of small objects and detailed regions. 

\subsection{Ablation Studies}
\subsubsection{Effectiveness of Different Modules}
To validate the effectiveness of each proposed module, we conduct ablation studies on the ScanNet and ICL-NUIM datasets. Five different configurations are investigated to evaluate the following modules: (1) Uncertainty-Guided Prior Filtering: applying the masked uncertainty learning to localize and filter out the noisy prior; (2) Uncertainty-Guided Ray sampling: sampling more rays from the localized uncertain regions; (3) Uncertainty-Guided Smooth: prohibiting the smooth regularization to the uncertain regions; (4) Bias-aware SDF to density transformation: utilizing curvature-based SDF mapping function to eliminate the bias in volume rendering. The quantitative results are shown in Table \ref{table:ablations}. The top row of Table \ref{table:ablations} corresponds to the baseline MonoSDF \cite{yu2022monosdf}.

Compared with MonoSDF \cite{yu2022monosdf}, our uncertainty-guided prior filtering improves the F-score by 3.53 and 6.58 on ScanNet \cite{dai2017scannet} and ICL-NUIM datasets, respectively. Based on this performance, the uncertainty-guided ray sampling and smooth regularization further improve the F-score by 1 to 3 on both datasets. Further, the final performance increases by 1.24 and 0.84 in terms of F-score on these two datasets by introducing the bias-aware transformation from SDF to density module. Among these proposed modules, the performance improvement benefiting from the uncertainty-guided prior filtering, uncertainty-guided ray sampling, and the transformation from SDF to density than that from the uncertainty-guided smooth regularization. 

\begin{table}[]
\caption{We evaluate the performance of our methods with different $\lambda$ in Eq. \ref{eq:importance}. We set the $\lambda$ as equal to 0.9.}
\centering
\resizebox{0.9\linewidth}{!}{
\begin{tabular}{c|ccccc}
\toprule
$\lambda$ & 1      & 0.9             & 0.5    & 0.1    & 0      \\ \hline
Chamfer$\downarrow$             & 0.095 & \textbf{0.093} & 0.095 & 0.096 & 0.097 \\
F-score$\uparrow$                & 77.16  & \textbf{77.39}  & 76.81  & 75.71  & 75.67  \\
Normal C$\uparrow$     & 88.19  & \textbf{88.34}  & 88.12  & 88.24  & 88.21     \\
\bottomrule
\end{tabular}
}
\label{table:lambda}
\end{table}

As shown in Fig. \ref{fig:ablation}, our method achieves a more stable reconstruction performance than MonoSDF. Some thin and detailed structures, such as the chair backrest and lamp, cannot be well reconstructed by MonoSDF, while these regions can be well reconstructed when applying prior filtering, uncertainty-guided ray sampling, and uncertainty-guided smooth regularization but still have structural damages. However, for those more tiny structures, such as the chair legs, still cannot be reconstructed, only applying these three modules is still not enough. But when introducing the bias-aware SDF to density transformation, these regions can be well reconstructed. 
This indicates that only detecting and filtering out the inaccurate geometry prior is not enough. A bias-aware SDF to density transformation is also indispensable for the reconstruction of the regions with finer details.

\begin{figure}[]
\centering
\includegraphics[width=0.43\textwidth]{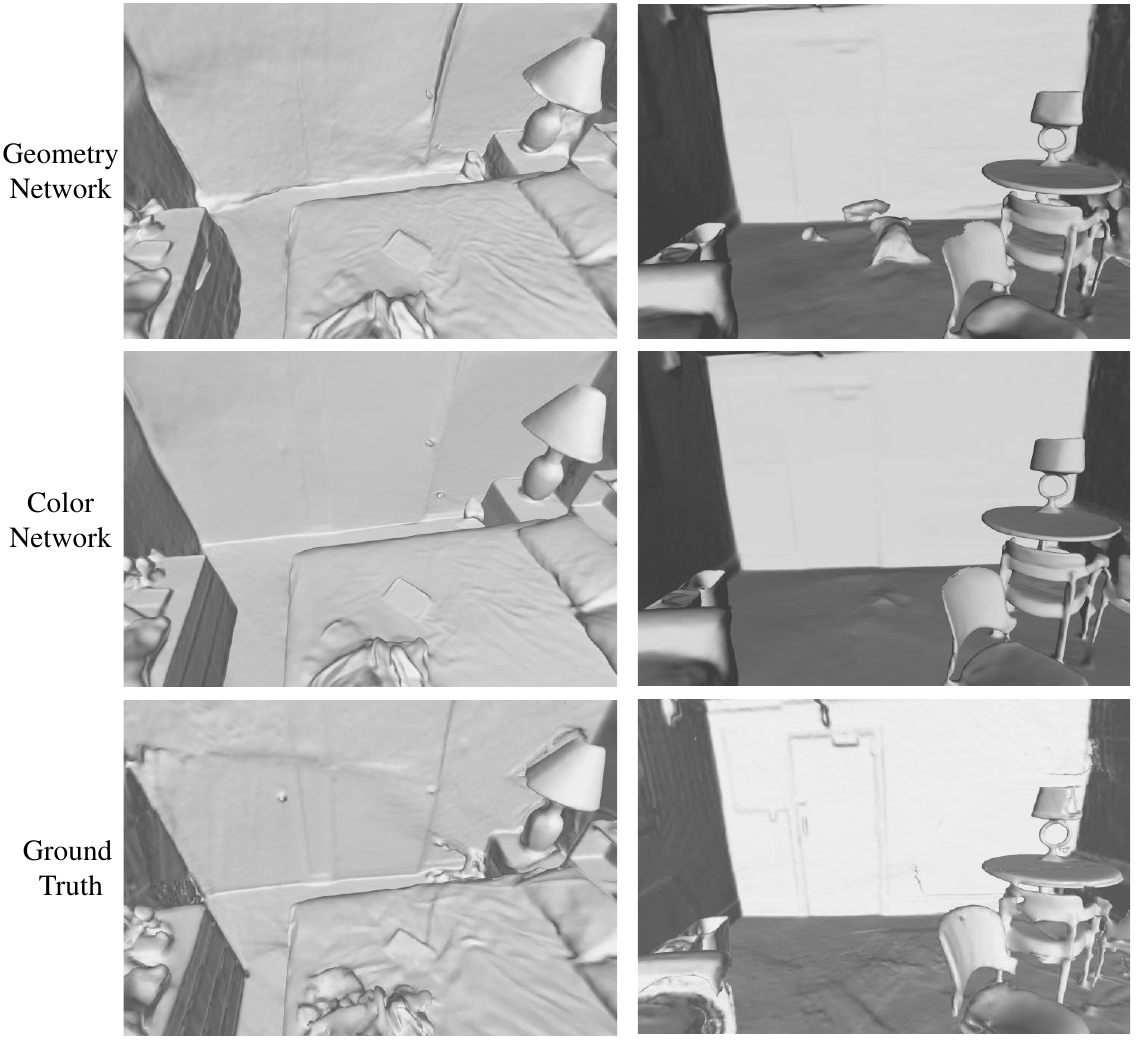}
\caption{
The "Geometry Network" indicates predicting the uncertainty by the geometry network, which models the uncertainty view independently. The "Color Network" indicates predicting the uncertainty by the color network which models the uncertainty view dependently.
}
\label{fig: geo_color}
\end{figure}

\begin{table}[t]
\centering
\caption{
We conduct experiments to evaluate the reconstruction performance of estimating the uncertainty by geometry network and color network on ScanNet \cite{dai2017scannet} dataset. Evaluating the uncertainty by the color network is better than the geometry network.}
\resizebox{\linewidth}{!}{
\begin{tabular}{c|cccccc}
\toprule
Method   & Acc$\downarrow$            & Comp$\downarrow$           & Pre$\uparrow$          & Recall$\uparrow$        & Chamfer$\downarrow$        & F-score $\uparrow$       \\ \hline
GeoNet   & 0.039          & 0.044          & 75.6          & 71.2          & 0.042          & 73.2          \\
ColorNet & \textbf{0.036} & \textbf{0.040} & \textbf{80.7} & \textbf{76.5} & \textbf{0.038} & \textbf{78.5} \\
\bottomrule
\end{tabular}
}
\label{tab: geo_color}
\end{table}

\begin{table*}[!h]
\caption{
The comparison experiments on the DTU dataset \cite{aanaes2016large}. We only modify the SDF to density transformation approach to the TUVR \cite{zhang2023towards} and Ours.
} 
\centering
\resizebox{18cm}{!}{
\begin{tabular}{c|cccccccccccccccc}
\toprule
Method & 24            & 37            & 40            & 55            & 63            & 65            & 69            & 83            & 97            & 105           & 106           & 110           & 114           & 118           & 122           & Mean          \\ \hline
NeuS\cite{wang2021neus}   & 1.00          & 1.37          & 0.93          & 0.43          & 1.10          & 0.65          & \textbf{0.57} & 1.48          & \textbf{1.09} & \textbf{0.83} & \textbf{0.52} & 1.2           & 0.35          & 0.49          & 0.54          & 0.84          \\
TUVR\cite{zhang2023towards}   & 0.83          & 1.06          & 0.57          & 0.40          & \textbf{1.00} & 0.62          & 0.62          & \textbf{1.41} & 1.32          & 0.94          & 0.57          & 1.07          & 0.35          & 0.49          & 0.51          & 0.79          \\
Ours   & \textbf{0.71} & \textbf{0.94} & \textbf{0.46} & \textbf{0.39} & 1.05          & \textbf{0.61} & 0.59          & 1.49          & 1.20          & 0.88          & 0.61          & \textbf{1.05} & \textbf{0.34} & \textbf{0.47} & \textbf{0.49} & \textbf{0.75}    \\
\bottomrule
\end{tabular}
}
\label{table:dtu}
\end{table*}

\begin{figure*}[!h]
\centering
\includegraphics[width=0.85\textwidth]{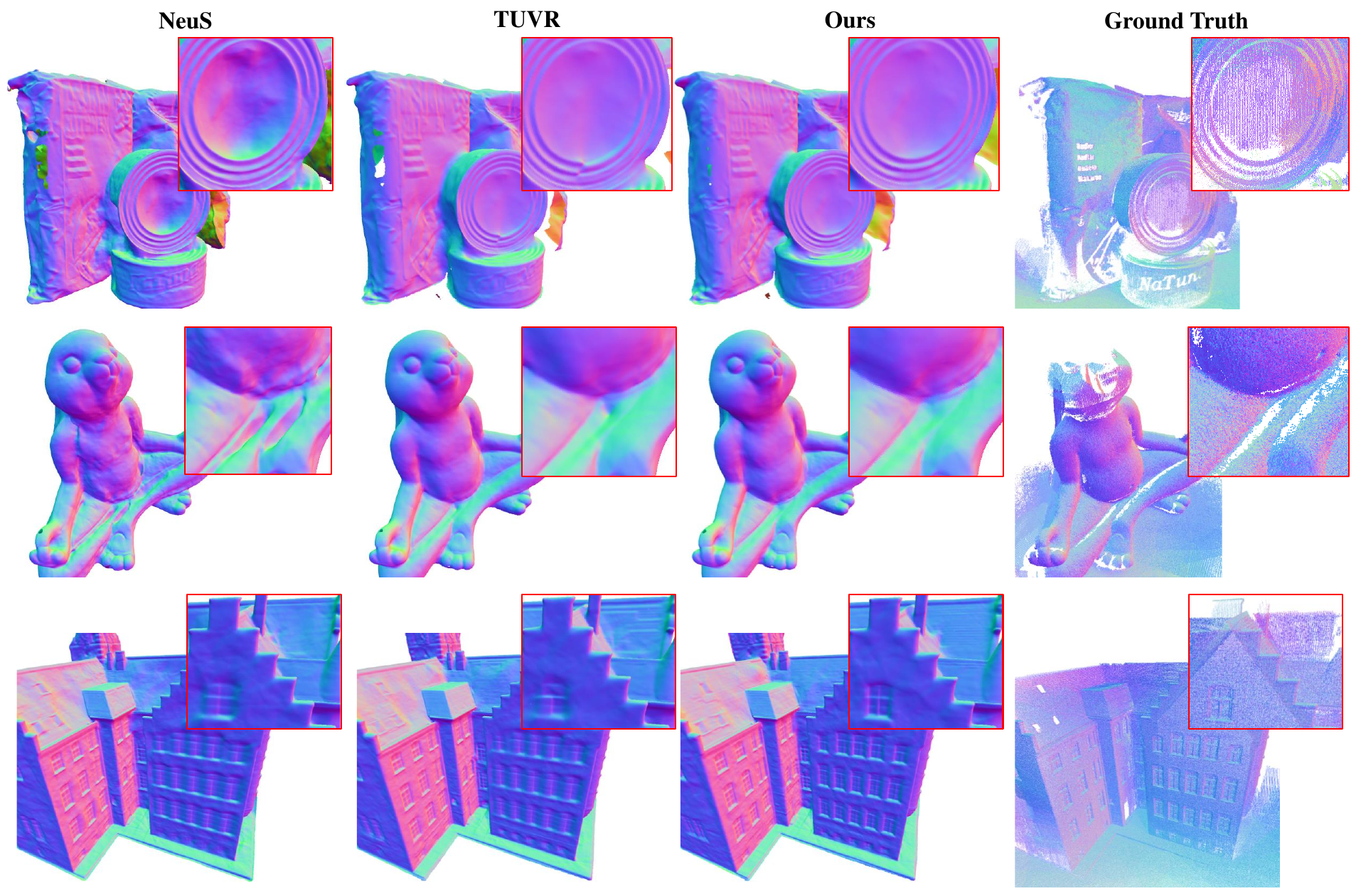}
\caption{
We conduct ablation studies on the DTU dataset \cite{aanaes2016large} to prove the effectiveness of the proposed SDF to density transformation. For our model and TUVR \cite{zhang2023towards}, we do not apply the geometry prior to reconstruction.}
\label{fig: dtu}
\end{figure*}

\subsubsection{Trade-off Between Different Uncertainties}
We introduce a hyper-parameter $\lambda$ in Eq. \ref{eq:importance} to fuse the localized results from the depth and normal prior. As shown in Table \ref{table:lambda}, we set the $\lambda$ with different values to study the effect of different priors in localizing the detailed and important regions. It can be observed that as $\lambda$ gradually decreases, the performance decreases. This indicates that applying the normal prior for uncertainty-guided prior filtering can lead to better performance than utilizing the depth prior. The reason is that the obtained depth prior from the pre-trained model is defined only up to scale, such that the optimization of uncertainty would be misled by the normalization of depth prior. The normal prior can provide more accurate information than the depth prior since it is not related to the distance scale problem. These experiments show that the normal prior is more important for reconstructing the details of the indoor scene. We set $\lambda=0.9$ in our implementation.

\subsubsection{Modeling the Uncertainty}
The uncertainty scores can be predicted by the geometry network or color network. The input of the geometry network is only the 3D space coordinate while the color network additionally considers the normal and ray direction. We conduct ablation studies to prove that modeling the prior uncertainty by the color network is better than a geometry network. The quantitative results are shown in Table \ref{tab: geo_color}. This proves that the modeling of uncertainty should be view-dependent. Suppose a point in 3D space can be viewed from different viewpoints, view-dependent modeling can predict different uncertainty scores for prior from different views such that the views with inaccurate prior can be filtered and the rest can be preserved.

More qualitative results are shown in Fig. \ref{fig: geo_color}. It can be observed that predicting the uncertainty score by the color network can reconstruct the scene more robustly than by the geometry network. The reason is that a region can be viewed from multiple viewpoints and only the viewpoints with inaccurate prior need to be filtered such that the uncertainty of region should be view-dependent.

\subsubsection{Efficiency of the SDF to Density Transformation}
To evaluate the efficiency of the proposed SDF to density transformation, we conduct the experiments by only applying this sub-module to the baseline on the DTU dataset \cite{aanaes2016large}. 

The results are shown in Table. \ref{table:dtu}. Our method achieves better performance than other baselines \cite{wang2021neus, zhang2023towards}. This proves the efficiency of the proposed SDF to density transformation approach. The qualitative results are shown in Fig. \ref{fig: dtu}, it can be observed that our method can reconstruct the detailed curved regions better than other baselines. These experiments prove that our proposed curvature-aware SDF to density transformation is efficient for the detailed and curved regions. Considering the curvature of the surface in the transformation of the SDF to the density can benefit the reconstruction because the bias is reduced.

\section{Conclusion}
We introduce DebSDF, which improved the detail and quality of indoor 3D reconstructions by localizing uncertainty regions and introducing a bias-aware SDF-to-density transformation for volume rendering of SDF. Based on the observation that a prior is correct if it is consistent with other priors, we propose an uncertainty modeling approach that effectively identifies large error regions in monocular geometric priors, which usually correspond to fine-detailed regions in the indoor scene. Accordingly, we selectively filter out geometry priors in these regions to avoid their potentially negative effect. We also assign a higher sampling probability to these regions and apply adaptive smooth regularization, further improving reconstruction quality. Furthermore, we found that the volume rendering technique of neural implicit surface used in previous work has a strong bias in eliminating fine-detailed surfaces. Consequently, we propose a progressively growing, bias-aware SDF-to-density transformation method to reduce the impact of these biases, enhancing the reconstruction of thin, detailed structures in indoor environments. Our DebSDF demonstrates improved reconstruction compared to previous work, evidenced by experiments across five challenging datasets.

\noindent\textbf{Limitation} While DebSDF improves the reconstruction quality significantly over previous work, several limitations still exist.
First, our method depends on the quality of monocular priors and could potentially benefit from future developments of monocular priors. Second, we use images and monocular priors with resolution $384\times 384$ because Omnidata\cite{eftekhar2021omnidata} model is trained on the images with this resolution, and their performances when using images with high resolution are poor. An alternative way is to resize the monocular priors to a high resolution, which we empirically found ineffective as it generates wrong priors to many pixels. More advanced methods using high-resolution priors \cite{yu2022monosdf} are left as our future work. 

\section{Acknowledge}
The work was supported by NSFC \# 62172279, \# 61932020, Program of Shanghai Academic Research Leader.
 
\bibliographystyle{IEEEtran}


\begin{thebibliography}{1}
\bibitem{dai2017scannet}
A.~Dai, A.~X. Chang, M.~Savva, M.~Halber, T.~Funkhouser, and M.~Nie{\ss}ner, ``Scannet: Richly-annotated 3d reconstructions of indoor scenes,'' in \emph{Proceedings of the IEEE/CVF Conference on Computer Vision and Pattern Recognition}, pp. 5828--5839, 2017.

\bibitem{handa2014benchmark}
A.~Handa, T.~Whelan, J.~McDonald, and A.~J. Davison, ``A benchmark for RGB-D visual odometry, 3D reconstruction and SLAM,'' in \emph{2014 IEEE International Conference on Robotics and Automation (ICRA)}, pp. 1524--1531, 2014, IEEE.

\bibitem{verbin2022ref}
D.~Verbin, P.~Hedman, B.~Mildenhall, T.~Zickler, J.~T. Barron, and P.~P. Srinivasan, ``Ref-nerf: Structured View-dependent Appearance for Neural Radiance Fields,'' in \emph{2022 IEEE/CVF Conference on Computer Vision and Pattern Recognition}, pp. 5481--5490, 2022, IEEE.

\bibitem{straub2019replica}
J.~Straub, T.~Whelan, L.~Ma, Y.~Chen, E.~Wijmans, S.~Green, J.~J. Engel, R.~Mur-Artal, C.~Ren, S.~Verma, \emph{et~al.}, ``The Replica dataset: A Digital Replica of Indoor Spaces,'' \emph{arXiv preprint arXiv:1906.05797}, 2019.

\bibitem{Knapitsch2017}
A. Knapitsch, J. Park, Q.-Y. Zhou, and V. Koltun, "Tanks and Temples: Benchmarking Large-Scale Scene Reconstruction," \emph{ACM Transactions on Graphics}, vol. 36, no. 4, 2017.

\bibitem{guo2022neural}
H. Guo, S. Peng, H. Lin, Q. Wang, G. Zhang, H. Bao, and X. Zhou, "Neural 3d scene reconstruction with the Manhattan-world assumption," in \emph{Proceedings of the IEEE/CVF Conference on Computer Vision and Pattern Recognition}, pp. 5511--5520, 2022.

\bibitem{yu2022monosdf}
Z. Yu, S. Peng, M. Niemeyer, T. Sattler, and A. Geiger, "MonoSDF: Exploring Monocular Geometric Cues for Neural Implicit Surface Reconstruction," in \emph{Advances in Neural Information Processing Systems}, 2022.

\bibitem{wang2022neuris}
J. Wang, P. Wang, X. Long, C. Theobalt, T. Komura, L. Liu, and W. Wang, "Neuris: Neural reconstruction of indoor scenes using normal priors," in \emph{Proceedings of the European Conference on Computer Vision}, 2022.

\bibitem{schonberger2016pixelwise}
J. L. Sch{\"o}nberger, E. Zheng, J.-M. Frahm, and M. Pollefeys, "Pixelwise view selection for unstructured multi-view stereo," in \emph{Proceedings of the European Conference on Computer Vision}, 2016.

\bibitem{yariv2021volume}
L. Yariv, J. Gu, Y. Kasten, and Y. Lipman, ``Volume Rendering of Neural Implicit Surfaces,'' \emph{Advances in Neural Information Processing Systems}, vol. 34, pp. 4805--4815, 2021.

\bibitem{wang2021neus}
P. Wang, L. Liu, Y. Liu, C. Theobalt, T. Komura, and W. Wang, ``NeuS: Learning Neural Implicit Surfaces by Volume Rendering for Multi-view Reconstruction,'' \emph{Advances in Neural Information Processing Systems}, vol. 34, pp. 27171--27183, 2021.

\bibitem{huang2018deepmvs}
P.-H. Huang, K. Matzen, J. Kopf, N. Ahuja, and J.-B. Huang, ``Deepmvs: Learning Multi-view Stereopsis,'' in \emph{Proceedings of the IEEE/CVF Conference on Computer Vision and Pattern Recognition}, pp. 2821--2830, 2018.

\bibitem{yao2018mvsnet}
Y. Yao, Z. Luo, S. Li, T. Fang, and L. Quan, ``Mvsnet: Depth inference for unstructured multi-view stereo,'' in \emph{Proceedings of the European Conference on Computer Vision}, pp. 767--783, 2018.

\bibitem{yao2019recurrent}
Y. Yao, Z. Luo, S. Li, T. Shen, T. Fang, and L. Quan, ``Recurrent MVSnet for High-Resolution Multi-View Stereo Depth Inference,'' in \emph{Proceedings of the IEEE/CVF Conference on Computer Vision and Pattern Recognition}, pp. 5525--5534, 2019.

\bibitem{cheng2020deep}
S. Cheng, Z. Xu, S. Zhu, Z. Li, L. E. Li, R. Ramamoorthi, and H. Su, ``Deep Stereo Using Adaptive Thin Volume Representation with Uncertainty Awareness,'' in \emph{Proceedings of the IEEE/CVF Conference on Computer Vision and Pattern Recognition}, pp. 2524--2534, 2020.

\bibitem{liao2021adaptive}
J. Liao, Y. Fu, Q. Yan, F. Luo, and C. Xiao, ``Adaptive Depth Estimation for Pyramid Multi-View Stereo,'' \emph{Computers \& Graphics}, vol. 97, pp. 268--278, Elsevier, 2021.

\bibitem{gu2020cascade}
X. Gu, Z. Fan, S. Zhu, Z. Dai, F. Tan, and P. Tan, ``Cascade Cost Volume for High-Resolution Multi-View stereo and Stereo Matching,'' in \emph{Proceedings of the IEEE/CVF Conference on Computer Vision and Pattern Recognition}, pp. 2495--2504, 2020.

\bibitem{yu2020fast}
Z. Yu and S. Gao, ``Fast-MVSNet: Sparse-to-Dense Multi-View Stereo With Learned Propagation and Gauss-Newton Refinement,'' in \emph{Proceedings of the IEEE/CVF Conference on Computer Vision and Pattern Recognition}, pp. 1949--1958, 2020.

\bibitem{wang2021patchmatchnet}
F. Wang, S. Galliani, C. Vogel, P. Speciale, and M. Pollefeys, ``PatchmatchNet: Learned Multi-View Patchmatch Stereo,'' in \emph{Proceedings of the IEEE/CVF Conference on Computer Vision and Pattern Recognition}, pp. 14194--14203, 2021.

\bibitem{ding2022transmvsnet}
Y. Ding, W. Yuan, Q. Zhu, H. Zhang, X. Liu, Y. Wang, and X. Liu, ``TransMVSNet: Global Context-aware Multi-view Stereo Network with Transformers,'' in \emph{Proceedings of the IEEE/CVF Conference on Computer Vision and Pattern Recognition}, pp. 8585--8594, 2022.

\bibitem{xu2021self}
H. Xu, Z. Zhou, Y. Qiao, W. Kang, and Q. Wu, ``Self-supervised Multi-view Stereo via Effective Co-Segmentation and Data-Augmentation,'' in \emph{Proceedings of the AAAI Conference on Artificial Intelligence}, vol. 35, no. 4, pp. 3030--3038, 2021.

\bibitem{bozic2021transformerfusion}
A. Bozic, P. Palafox, J. Thies, A. Dai, and M. Nie{\ss}ner, ``TransformerFusion: Monocular RGB Scene Reconstruction using Transformers,'' \emph{Advances in Neural Information Processing Systems}, vol. 34, pp. 1403--1414, 2021.

\bibitem{oechsle2021unisurf}
M. Oechsle, S. Peng, and A. Geiger, ``UNISURF: Unifying Neural Implicit Surfaces and Radiance Fields for Multi-View Reconstruction,'' in \emph{Proceedings of the IEEE/CVF International Conference on Computer Vision}, pp. 5589--5599, 2021.

\bibitem{mildenhall2021nerf}
B. Mildenhall, P. P. Srinivasan, M. Tancik, J. T. Barron, R. Ramamoorthi, and R. Ng, ``Nerf: Representing Scenes as Neural Radiance Fields for View Synthesis,'' \emph{Communications of the ACM}, vol. 65, no. 1, pp. 99--106, ACM New York, NY, USA, 2021.

\bibitem{mescheder2019occupancy}
L. Mescheder, M. Oechsle, M. Niemeyer, S. Nowozin, and A. Geiger, ``Occupancy Networks: Learning 3D Reconstruction in Function Space,'' in \emph{Proceedings of the IEEE/CVF Conference on Computer Vision and Pattern Recognition}, pp. 4460--4470, 2019.

\bibitem{park2019deepsdf}
J. J. Park, P. Florence, J. Straub, R. Newcombe, and S. Lovegrove, ``DeepSDF: Learning Continuous Signed Distance Functions for Shape Representation,'' in \emph{Proceedings of the IEEE/CVF Conference on Computer Vision and Pattern Recognition}, pp. 165--174, 2019.

\bibitem{wei2021nerfingmvs}
Y. Wei, S. Liu, Y. Rao, W. Zhao, J. Lu, and J. Zhou, ``NerfingMVS: Guided Optimization of Neural Radiance Fields for Indoor Multi-view Stereo,'' in \emph{Proceedings of the IEEE/CVF International Conference on Computer Vision}, pp. 5610--5619, 2021.

\bibitem{yariv2020multiview}
L. Yariv, Y. Kasten, D. Moran, M. Galun, M. Atzmon, B. Ronen, and Y. Lipman, ``Multiview Neural Surface Reconstruction by Disentangling Geometry and Appearance,'' \emph{Advances in Neural Information Processing Systems}, vol. 33, pp. 2492--2502, 2020.

\bibitem{barron2021mip}
J. T. Barron, B. Mildenhall, M. Tancik, P. Hedman, R. Martin-Brualla, and P. P. Srinivasan, ``Mip-NeRF: A Multiscale Representation for Anti-Aliasing Neural Radiance Fields,'' in \emph{Proceedings of the IEEE/CVF International Conference on Computer Vision}, pp. 5855--5864, 2021.

\bibitem{murez2020atlas}
Z.~Murez, T.~Van As, J.~Bartolozzi, A.~Sinha, V.~Badrinarayanan, and A.~Rabinovich, ``Atlas: End-to-end 3d Scene Reconstruction from Posed Images,'' in \emph{Proceedings of the European Conference on Computer Vision}, pp. 414--431, Springer, 2020.

\bibitem{sun2021neuralrecon}
J.~Sun, Y.~Xie, L.~Chen, X.~Zhou, and H.~Bao, ``NeuralRecon: Real-Time Coherent 3D Reconstruction from Monocular Video,'' in \emph{Proceedings of the IEEE/CVF Conference on Computer Vision and Pattern Recognition}, pp. 15598--15607, 2021.

\bibitem{roessle2022dense}
B.~Roessle, J.~T. Barron, B.~Mildenhall, P.~P. Srinivasan, and M.~Nie{\ss}ner, ``Dense Depth Priors for Neural Radiance Fields from Sparse Input Views,'' in \emph{Proceedings of the IEEE/CVF Conference on Computer Vision and Pattern Recognition}, pp. 12892--12901, 2022.

\bibitem{deng2022depth}
K.~Deng, A.~Liu, J.-Y. Zhu, and D.~Ramanan, ``Depth-supervised NeRF: Fewer Views and Faster Training for Free'', in \emph{Proceedings of the IEEE/CVF Conference on Computer Vision and Pattern Recognition}, pp. 12882--12891, 2022.

\bibitem{eftekhar2021omnidata}
A.~Eftekhar, A.~Sax, J.~Malik, and A.~Zamir, ``Omnidata: A Scalable Pipeline for Making Multi-Task Mid-Level Vision Datasets from 3D Scans,'' in \emph{Proceedings of the IEEE/CVF International Conference on Computer Vision}, pp. 10786--10796, 2021.

\bibitem{coughlan1999manhattan}
J.~M. Coughlan and A.~L. Yuille, ``Manhattan World: Compass Direction from a Single Image by Bayesian Inference,'' in \emph{Proceedings of the IEEE/CVF International Conference on Computer Vision}, vol. 2, pp. 941--947, IEEE, 1999.

\bibitem{newcombe2011kinectfusion}
R.~A. Newcombe, S.~Izadi, O.~Hilliges, D.~Molyneaux, D.~Kim, A.~J. Davison, P.~Kohi, J.~Shotton, S.~Hodges, and A.~Fitzgibbon, ``Kinectfusion: Real-time dense surface mapping and tracking,'' in \emph{2011 10th IEEE International Symposium on Mixed and Augmented Reality}, pp. 127--136, IEEE, 2011.

\bibitem{schonberger2016structure}
J.~L. Schonberger and J.-M. Frahm, Structure-from-motion Revisited, in \emph{Proceedings of the IEEE/CVF Conference on Computer Vision and Pattern Recognition}, pp. 4104--4113, 2016.

\bibitem{merrell2007real}
P.~Merrell, A.~Akbarzadeh, L.~Wang, P.~Mordohai, J.-M. Frahm, R.~Yang, D.~Nist{\'e}r, and M.~Pollefeys, ``Real-time visibility-based fusion of depth maps,'' in \emph{Proceedings of the IEEE /CVF International Conference on Computer Vision}, pp. 1--8, IEEE, 2007.

\bibitem{bleyer2011patchmatch}
M.~Bleyer, C.~Rhemann, and C.~Rother, ``Patchmatch stereo-stereo matching with slanted support windows,'' in \emph{The British Machine Vision Conference}, vol. 11, pp. 1--11, 2011.

\bibitem{paschalidou2018raynet}
D.~Paschalidou, O.~Ulusoy, C.~Schmitt, L.~Van Gool, and A.~Geiger, ``RRayNet: Learning Volumetric 3D Reconstruction with Ray Potentials,'' in \emph{Proceedings of the IEEE/CVF Conference on Computer Vision and Pattern Recognition}, 2018, pp. 3897--3906.

\bibitem{seitz1999photorealistic}
S.~M. Seitz and C.~R. Dyer, ``Photorealistic Scene Reconstruction by Voxel Coloring,'' \emph{International Journal of Computer Vision}, vol. 35, pp. 151--173, 1999, Springer.

\bibitem{ulusoy2015towards}
A.~O. Ulusoy, A.~Geiger, and M.~J. Black, ``Towards Probabilistic Volumetric Reconstruction Using Ray Potentials,'' in \emph{International Conference on 3D Vision}, IEEE, 2015, pp. 10--18.

\bibitem{niemeyer2022regnerf}
M.~Niemeyer, J.~T. Barron, B.~Mildenhall, M.~S.~M. Sajjadi, A.~Geiger, and N.~Radwan, ``RegNeRF: Regularizing Neural Radiance Fields for View Synthesis from Sparse Inputs,'' in \emph{Proceedings of the IEEE/CVF Conference on Computer Vision and Pattern Recognition}, 2022, pp. 5480--5490.

\bibitem{zhi2021place}
S.~Zhi, T.~Laidlow, S.~Leutenegger, and A.~J. Davison, ``In-place scene labelling and understanding with implicit scene representation,'' in \emph{Proceedings of the IEEE/CVF International Conference on Computer Vision}, 2021, pp. 15838--15847.

\bibitem{jain2021putting}
A.~Jain, M.~Tancik, and P.~Abbeel, ``Putting NeRF on a Diet: Semantically Consistent Few-Shot View Synthesis,'' in \emph{Proceedings of the IEEE/CVF International Conference on Computer Vision}, 2021, pp. 5885--5894.

\bibitem{eigen2014depth}
D.~Eigen, C.~Puhrsch, and R.~Fergus, ``Depth Map Prediction from a Single Image Using a Multi-Scale Deep Network,'' \emph{Advances in Neural Information Processing Systems}, vol. 27, 2014.

\bibitem{niemeyer2020differentiable}
M.~Niemeyer, L.~Mescheder, M.~Oechsle, and A.~Geiger, ``Differentiable Volumetric Rendering: Learning Implicit 3D Representations without 3D Supervision,'' in \emph{Proceedings of the IEEE/CVF Conference on Computer Vision and Pattern Recognition}, 2020, pp. 3504--3515.

\bibitem{gropp2020implicit}
A.~Gropp, L.~Yariv, N.~Haim, M.~Atzmon, and Y.~Lipman, ``Implicit Geometric Regularization for Learning Shapes,'' in \emph{Proceedings of the 37th International Conference on Machine Learning}, 2020, pp. 3789--3799.

\bibitem{novello2022exploring}
T.~Novello, G.~Schardong, L.~Schirmer, V.~da Silva, H.~Lopes, and L.~Velho, ``Exploring Differential Geometry in Neural Implicits,'' \emph{Computers \& Graphics}, vol. 108, pp. 49--60, 2022, Elsevier.

\bibitem{wujun2007lines}
W. Che, J.-C. Paul, and X. Zhang, ``Lines of curvature and umbilical points for implicit surfaces,'' \emph{Computer Aided Geometric Design}, pp. 395--409, 2007, Elsevier.

\bibitem{kazhdan2013screened}
M. Kazhdan and H. Hoppe, ``Screened Poisson Surface Reconstruction,'' \emph{ACM Transactions on Graphics}, vol. 32, no. 3, pp. 1--13, 2013, ACM New York, NY, USA.

\bibitem{shen2022conditional}
J. Shen, A. Agudo, F. Moreno-Noguer, and A. Ruiz, ``Conditional-flow NeRF: Accurate 3D Modelling with Reliable Uncertainty Quantification,'' in \emph{Proceedings of the European Conference on Computer Vision}, Springer, 2022, pp. 540--557.

\bibitem{pan2022activenerf}
X. Pan, Z. Lai, S. Song, and G. Huang, ``ActiveNeRF: Learning Where to See with Uncertainty Estimation,'' in \emph{Proceedings of the European Conference on Computer Vision}, Springer, 2022, pp. 230--246.

\bibitem{antoran2020depth}
J. Antorán, J. Allingham, and J. M. Hernández-Lobato, ``Depth Uncertainty in Neural Networks,'' \emph{Advances in Neural Information Processing Systems}, vol. 33, pp. 10620--10634, 2020.

\bibitem{bae2021estimating}
G.~Bae, I.~Budvytis, and R.~Cipolla, ``Estimating and Exploiting the Aleatoric Uncertainty in Surface Normal Estimation,'' in \emph{Proceedings of the IEEE/CVF International Conference on Computer Vision}, 2021, pp. 13137--13146.

\bibitem{paszke2019pytorch}
A.~Paszke, S.~Gross, F.~Massa, A.~Lerer, J.~Bradbury, G.~Chanan, T.~Killeen, Z.~Lin, N.~Gimelshein, L.~Antiga, \textit{et al.}, ``PyTorch: An Imperative Style, High-Performance Deep Learning Library,'' \emph{Advances in Neural Information Processing Systems}, vol. 32, 2019.

\bibitem{muller2022instant}
T.~M{\"u}ller, A.~Evans, C.~Schied, and A.~Keller, ``Instant Neural Graphics Primitives with a Multiresolution Hash Encoding,'' \emph{ACM Transactions on Graphics}, vol. 41, no. 4, pp. 1--15, 2022.

\bibitem{tosi2021smd}
F.~Tosi, Y.~Liao, C.~Schmitt, and A.~Geiger, ``SMD-Nets: Stereo Mixture Density Networks,'' in \emph{Proceedings of the IEEE/CVF Conference on Computer Vision and Pattern Recognition}, 2021.

\bibitem{zhang2023towards}
Y.~Zhang, Z.~Hu, H.~Wu, M.~Zhao, L.~Li, Z.~Zou, and C.~Fan, ``Towards Unbiased Volume Rendering of Neural Implicit Surfaces with Geometry Priors,'' in \emph{Proceedings of the IEEE/CVF Conference on Computer Vision and Pattern Recognition}, 2023, pp. 4359--4368.

\bibitem{barron2022mip}
J.~T. Barron, B.~Mildenhall, D.~Verbin, P.~P. Srinivasan, and P.~Hedman, ``Mip-NeRF 360: Unbounded Anti-Aliased Neural Radiance Fields,'' in \emph{Proceedings of the IEEE/CVF Conference on Computer Vision and Pattern Recognition}, 2022, pp. 5470--5479.

\bibitem{lorensen1987marching}
W.~E. Lorensen and H.~E. Cline, ``Marching cubes: A high resolution 3D surface construction algorithm,'' \emph{ACM SIGGRAPH}, vol. 21, no. 4, pp. 163--169, 1987, ACM New York, NY, USA.

\bibitem{wang2022hf}
Y.~Wang, I.~Skorokhodov, and P.~Wonka, ``HF-NeuS: Improved Surface Reconstruction Using High-Frequency Details,'' \emph{Advances in Neural Information Processing Systems}, vol. 35, pp. 1966--1978, 2022.

\bibitem{dong2023fast}
W.~Dong, C.~Choy, C.~Loop, O.~Litany, Y.~Zhu, and A.~Anandkumar, ``Fast Monocular Scene Reconstruction with Global-Sparse Local-Dense Grids,'' in \emph{Proceedings of the IEEE/CVF Conference on Computer Vision and Pattern Recognition}, 2023, pp. 4263--4272.

\bibitem{liang2023helixsurf}
Z.~Liang, Z.~Huang, C.~Ding, and K.~Jia, ``HelixSurf: A Robust and Efficient Neural Implicit Surface Learning of Indoor Scenes with Iterative Intertwined Regularization,'' in \emph{Proceedings of the IEEE/CVF Conference on Computer Vision and Pattern Recognition}, 2023, pp. 13165--13174.

\bibitem{ye2023self}
B.~Ye, S.~Liu, X.~Li, and M.-H. Yang, ``Self-Supervised Super-Plane for Neural 3D Reconstruction,'' in \emph{Proceedings of the IEEE/CVF Conference on Computer Vision and Pattern Recognition}, 2023, pp. 21415--21424.

\bibitem{li2023neuralangelo}
Z.~Li, T.~M{\"u}ller, A.~Evans, R.~H. Taylor, M.~Unberath, M.-Y. Liu, and C.-H. Lin, ``Neuralangelo: High-Fidelity Neural Surface Reconstruction,'' in \emph{Proceedings of the IEEE/CVF Conference on Computer Vision and Pattern Recognition}, 2023, pp. 8456--8465.

\bibitem{Yu2022SDFStudio}
Z.~Yu, A.~Chen, B.~Antic, S.~Peng, A.~Bhattacharyya, M.~Niemeyer, S.~Tang, T.~Sattler, and A.~Geiger, ``SDFStudio: A Unified Framework for Surface Reconstruction,'' 2022, \url{https://github.com/autonomousvision/sdfstudio}.

\bibitem{nerfstudio}
M.~Tancik, E.~Weber, E.~Ng, R.~Li, B.~Yi, J.~Kerr, T.~Wang, A.~Kristoffersen, J.~Austin, K.~Salahi, A.~Ahuja, D.~McAllister, and A.~Kanazawa, ``Nerfstudio: A Modular Framework for Neural Radiance Field Development,'' in \emph{ACM SIGGRAPH}, 2023.

\bibitem{qu2021bayesian}
C.~Qu, W.~Liu, and C.~J. Taylor, ``Bayesian Deep Basis Fitting for Depth Completion with Uncertainty,'' in \emph{Proceedings of the IEEE/CVF International Conference on Computer Vision}, 2021, pp. 16147--16157.

\bibitem{eldesokey2020uncertainty}
A.~Eldesokey, M.~Felsberg, K.~Holmquist, and M.~Persson, ``Uncertainty-aware CNNs for Depth Completion: Uncertainty from Beginning to End,'' in \emph{Proceedings of the IEEE/CVF Conference on Computer Vision and Pattern Recognition}, 2020, pp. 12014--12023.

\bibitem{aanaes2016large}
H.~Aan{\ae}s, R.~R. Jensen, G.~Vogiatzis, E.~Tola, and A.~B. Dahl, ``Large-scale Data for Multiple-view Stereopsis,'' \emph{International Journal of Computer Vision}, vol. 120, pp. 153--168, 2016, Springer.

\end{thebibliography}


\end{document}